\documentclass[10pt,twocolumn,letterpaper]{article}

\usepackage[pagenumbers]{iccv} 

\definecolor{iccvblue}{rgb}{0.21,0.49,0.74}
\usepackage[pagebackref,breaklinks,colorlinks,allcolors=iccvblue]{hyperref}
\usepackage{multirow}
\usepackage[normalem]{ulem}
\usepackage{makecell}
\usepackage{array}
\usepackage{amsmath}
\usepackage{graphicx}
\usepackage{amssymb}
\usepackage{booktabs}
\usepackage{colortbl}

\newcolumntype{L}[1]{>{\raggedright\let\newline\\\arraybackslash\hspace{0pt}}m{#1}}
\newcolumntype{C}[1]{>{\centering\let\newline\\\arraybackslash\hspace{0pt}}m{#1}}
\newcolumntype{R}[1]{>{\raggedleft\let\newline\\\arraybackslash\hspace{0pt}}m{#1}}
\newcolumntype{P}[1]{>{\centering\arraybackslash}p{#1}}
\newcommand{\overbar}[1]{\mkern 1.5mu\overline{\mkern-1.5mu#1\mkern-1.5mu}\mkern 1.5mu}
\usepackage{tabularx}
\newcolumntype{Z}{>{\centering\arraybackslash}X}

\def\paperID{323} 
\def\confName{ICCV}
\def\confYear{2025}

\title{VIGFace: Virtual Identity Generation for Privacy-Free Face Recognition}

\author{
  Minsoo Kim$^*$ \\
  Korea Institute of Sciene and Technology\\
  Seoul, Korea \\
  {\tt\small kim1102@kist.re.kr}
  \and
  Min-Cheol Sagong$^*$ \\
  Korea Institute of Sciene and Technology \\
  Seoul, Korea \\
  {\tt\small mcsagong@kist.re.kr}
  \and
  Gi Pyo Nam \\
  Korea Institute of Sciene and Technology\\
  Seoul, Korea \\
  {\tt\small gpnam@kist.re.kr}
  \and
  Junghyun Cho \\
  Korea Institute of Sciene and Technology\\
  Seoul, Korea \\
  {\tt\small jhcho@kist.re.kr}
  \and
  Ig-Jae Kim \\
  Korea Institute of Sciene and Tech.\\
  Seoul, Korea \\
  {\tt\small drjay@kist.re.kr}
}

\begin{document}
\maketitle

\begin{abstract}
Deep learning-based face recognition continues to face challenges due to its reliance on huge datasets obtained from web crawling, which can be costly to gather and raise significant real-world privacy concerns. To address this issue, we propose \textbf{VIGFace}, a novel framework capable of generating synthetic facial images. Our idea originates from pre-assigning virtual identities in the feature space. Initially, we train the face recognition model using a real face dataset and create a feature space for both real and virtual identities, where virtual prototypes are orthogonal to other prototypes. 
Subsequently, we train the diffusion model based on the established feature space, enabling it to generate authentic human face images from real prototypes and synthesize virtual face images from virtual prototypes.
Our proposed framework provides two significant benefits. 
Firstly, it shows clear separability between existing individuals and virtual face images, allowing one to create synthetic images with confidence and without concerns about privacy and portrait rights. Secondly, it ensures improved performance through data augmentation by incorporating real existing images. Extensive experiments demonstrate the superiority of our virtual face dataset and framework, outperforming the previous state-of-the-art on various face recognition benchmarks.
\href{https://github.com/kim1102/VIGFace}{https://github.com/kim1102/VIGFace}
\end{abstract}
\begin{figure*}[!ht]
\center
\includegraphics[width=\linewidth]{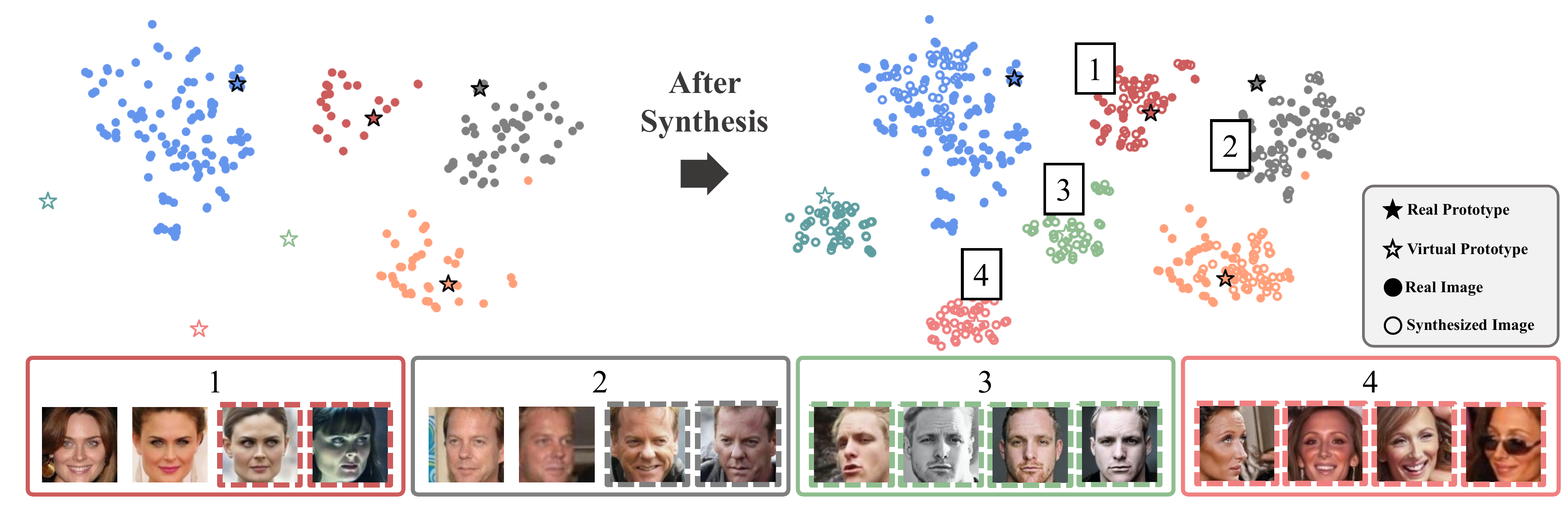}
\caption{T-distributed Stochastic Neighbor Embedding (T-SNE) \cite{van2008visualizing} plot of embeddings from real and synthesized images. The filled and lined stars represent the real and virtual prototypes, while filled and lined circles indicate the embeddings of real and synthesized images, respectively. The bottom of the figure shows the face images included in the cluster, and the dotted outlined images represent the face images generated using our method.}
\label{tsne}
\end{figure*}

\section{Introduction}
\label{sec:intro}

Deep learning-based Face Recognition (FR) models have significantly improved their performance due to recent advances in network architectures~\cite{simonyan2014very, srivastava2014dropout, he2015delving, ioffe2015batch, he2016deep, hu2018squeeze} and enhancements in loss functions~\cite{schroff2015facenet, sankaranarayanan2016triplet, wen2016discriminative, liu2017sphereface, deng2019arcface, kim2022adaface}.
The latest FR models utilize the softmax-variant loss for training to reduce intra-class variance and increase inter-class separability in the embedding space. This necessitates a very large dataset with numerous distinct individuals, significant variations within each individual for intra-class variance, and precise labels of subject identities (IDs). However, datasets are typically collected through web-crawling and then refined using automatic techniques that employ FR logits~\cite{deng2020sub, zhu2021webface260m}. Although this approach is successful in eliminating mislabeled data, it still struggles with persistent issues of small intra-class variance. 
Moreover, in the face recognition field, unlike other ordinary image classifications, there are practical issues such as portrait rights, making it even more difficult to collect training data.
For example, data sets such as those referenced in~\cite{yi2014learning, guo2016ms, zhu2021webface260m} consist of images of celebrities collected from the internet without consent. Furthermore, the datasets mentioned in~\cite{2016megaface, nech2017level} include facial images of the general population, including children. Due to privacy concerns, public access to these datasets has been revoked~\cite{van2020ethical}.

Synthetic datasets have been used to address the limitations caused by the scarcity of real datasets~\cite{tremblay2018training, eldersim, bsxgan} and biases present in the available real datasets~\cite{van2021decaf, wearmask3d}.
In the realm of face recognition, artificial faces show promise in addressing the aforementioned issues related to real face datasets. Generated faces are at low risk for label noise due to conditional generation. Bias problems, such as long-tailed distributions, which lead to class imbalances, can be mitigated by data augmentation. Importantly, there are no privacy concerns with virtual identities facial images if the method is effective.
Therefore, when creating artificial datasets, it is essential to consider: 1) ensuring that the generated data mirror the real data distribution, 2) the ability to generate new subjects separated from real data, and 3) maintaining ID consistency for each subject.

Previous attempts to create artificial face datasets have addressed some of the three aspects individually, but to the best of our knowledge, none have simultaneously taken into account all three aspects\cite{bae2023digiface, qiu2021synface, kim2023dcface, boutros2023idiff}. SynFace~\cite{qiu2021synface} introduces a high-quality synthetic face image dataset, closely reminiscent of real face images, using DiscoFaceGAN~\cite{deng2020disentangled}. However, DiscoFaceGAN is only capable of generating a limited number of unique subjects, fewer than 500~\cite{kim2023dcface}. On the other hand, DigiFace~\cite{bae2023digiface} utilizes 3D parametric modeling to create synthetic face images of various subjects. However, it faces difficulties in accurately replicating the quality and style distribution observed in real face images. DCFace~\cite{kim2023dcface} proposes a diffusion-based method to generate data that maintains the style of real datasets and ensures label consistency. However, DCFace primarily focuses on creating synthetic data rather than enhancing the capabilities of data augmentation. IDiffFace~\cite{boutros2023idiff} utilizes identity-conditioned latent diffusion to synthesize facial images from FR feature representations. However, it fails to guarantee the uniqueness of the synthesized identities. CemiFace~\cite{sun2025cemiface} and HSFace~\cite{wu2024vec2face} demonstrate high performance on the various FR benchmarks. However, we found identity leakage of the training dataset from both CemiFace and HSFace, which is critical from a privacy standpoint.

\begin{figure*}[t!]
\center
\includegraphics[width=1.0\linewidth]{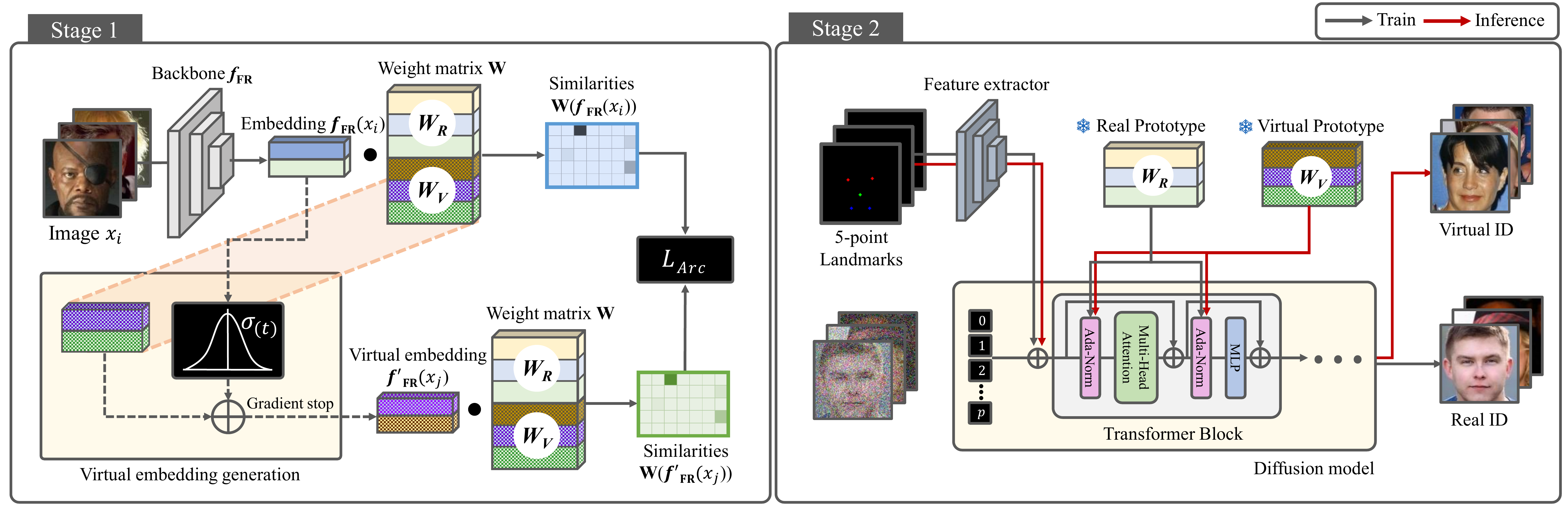}
\caption{
Pipeline for the proposed method. Conventional FR training includes prototypes for only real individuals, indicated as $W_{R}$. We add $k$ prototypes for virtual IDs, denoted as $W_{V}$. The virtual embedding $f'_{\mathrm{FR}}(x_{j})$ corresponding to the virtual person ID: $j$ is generated to follow distribution of the real embeddings. To synthesize the facial image from virtual prototypes, we adopt the DiT architecture~\cite{peebles2023scalable}, following the design approach of the Vision Transformer~(ViT)~\cite{dosovitskiy2010image}. Additionally, we adjust the DiT model to utilize 5-point landmark images to handle pose variations.}
\label{fig:pipeline}
\end{figure*}

The virtual data generated by our method possesses all three essential properties of virtual data mentioned above. The main concept of our paper is to incorporate virtual prototypes into the FR model. Virtual prototypes are trained simultaneously with real prototypes so that they pre-assign on the same feature space. The diffusion model is trained to preserve the identity of the individuals when generating face images based on the FR embedding. As virtual prototypes are allocated within the same embedding space as real prototypes, the produced face images are ensured to reflect the true data distribution. In addition, since prototypes are orthogonal to others, virtual subjects are guaranteed to be distinguished from the original data.~\cref{tsne} shows a toy example that visualizes the embeddings of real and virtual subjects to demonstrate the effectiveness of the suggested method. The virtual embedding can be distinct from the real individual clusters, whereas the images generated from virtual prototypes form unique clusters. The visualization demonstrates that our framework generates distinctive virtual human faces with high consistency in ID. The proposed method effectively addresses privacy concerns by creating datasets of non-existent individuals and achieving state-of-the-art performance compared to models trained with previous virtual face generation methods. Additionally, the FR model, which was trained using a combination of real and synthetic images together, achieves better performance than the model trained using only real images. This shows that the proposed method is also superior from the perspective of a data augmentation method.

Our contribution can be summarized as follows:
\begin{itemize}
\item[$\bullet$] Introducing VIGFace, a method designed to generate virtual identities with realistic appearance, guided by three main criteria: generating novel subjects that are not found in the real-world, consistently preserving the identity, but ensuring diverse characteristics for each individual.
\item[$\bullet$] Showing that the synthetic data generated by VIGFace can leverage intra-class diversity and inter-class variance, achieving SOTA performance in face recognition.
\item[$\bullet$] Releasing the virtual-only face dataset that can fully substitute the real dataset, helping alleviate privacy concerns.
\end{itemize}
\section{Related work}

\subsection{Face Recognition Models}
Current state-of-the-arts~(SOTA) FR methods, such as~\cite{liu2017sphereface, wang2018cosface, deng2019arcface, huang2020curricularface, meng2021magface, deng2021variational, boutros2022elasticface, kim2022adaface}, are designed based on softmax loss, but particularly utilize the angular/cosine distance instead of the Euclidean distance. The goal of these methods is to maximize the similarity between the embeddings and the ground-truth (GT) prototype, while minimizing the similarity between the embedding and prototypes of other classes. As a result, all embeddings in the same class, including prototypes, converge while maintaining their maximum distance from other classes. Therefore, if the embedding dimension is large enough, the clusters of each class will be nearly orthogonal \cite{deng2019arcface, brauchart2018random}. Naturally, a larger dataset that contains a greater variety of images allows for the training of better-performing FR models. Therefore, over time, larger and more diverse datasets~\cite{huang2012learning, guo2016ms, an2021partial, zhu2021webface260m} have been collected and published in the academic field. However, issues such as portrait rights and the high cost of building large-scale data remain unresolved.

\subsection{Synthetic Face Image Generation}
Synthetic training datasets have shown advantages in areas such as face recognition~\cite{bae2023digiface, qiu2021synface} and anti-spoofing\cite{liu2022spoof, sun2023rethinking}. 
They can address issues such as class imbalance, informed consent, and racial bias, which are often found in conventional large-scale face datasets~\cite{deng2019arcface, qiu2021synface, bae2023digiface}. Additionally, synthetic datasets support the importance of ethical considerations, such as portrait rights. Despite their benefits, the use of synthetic datasets lacks widespread acceptance, causing reduced recognition accuracy when used as the only training data.

SynFace~\cite{qiu2021synface} proposed a method that uses Disco-FaceGAN\cite{deng2020disentangled} to synthesize virtual faces. DigiFace-1M~\cite{bae2023digiface} suggests 3D model-based face rendering with image augmentations to create virtual face data and showed that the FR model could be trained using only virtual face images. DCFace~\cite{kim2023dcface} proposes a diffusion-based face generator combining subject appearance~(ID) and external factor~(style) conditions. IDiffFace~\cite{boutros2023idiff} uses latent diffusion conditioned with FR identity to generate a synthetic image from FR feature representation. Furthermore, GanDiffFace~\cite{melzi2023gandiffface} uses GANs to generate identity features. Vec2Face~\cite{wu2024vec2face} introduces a masked autoencoder to control the identity of face images and their attributes. CemiFace~\cite{sun2025cemiface} produces facial samples with various levels of similarity to the center of the subject. Although the above methods have succeeded in generating separable identities, they still need additional networks or thresholding to sample synthetic identities.

\section{Methods}
Our framework comprises two stages. First, we train the FR model using the real face dataset and design the feature space for both real and virtual IDs. Second, synthetic images are generated using the diffusion model based on the feature space of the pre-trained FR model that was trained in the first stage. This section provides a detailed explanation of each stage of the proposed framework.

\subsection{Stage 1: FR Model Training}

The proposed framework begins with training the FR model using real face images. This stage serves two purposes: 1) Training the FR model to achieve prototype features from face images, which are necessary for training the diffusion model in the second stage, and 2) Simultaneously assigning the positions of both the real ID and the virtual ID on the feature space. We choose ArcFace~\cite{deng2019arcface} to train the FR model in this stage. The ArcFace loss used to train the FR backbone $f_{\mathrm{FR}}$ and the prototype $W = [w_{1}, w_{2}, ..., w_{n}]$ can be described as follows:
\begin{equation}
L_{\mathrm{arc}} = -\log {\frac{e^{s\,\cos(\theta_{y_{i}}+m)}}{e^{s\,\cos(\theta_{y_{i}}+m)}+\sum_{j=1,j\neq y_{i}}^{n} e^{s\, \cos \theta_{j}}}},
\label{angular_arc_loss}
\end{equation}
\begin{equation}
    \cos \theta_{j} = \frac{{w_j}^T f_{\mathrm{FR}}(x_{i})}{\| w_{j} \| \| f_{\mathrm{FR}}(x_{i}) \|},
\end{equation}
where $n$ denotes the total number of real IDs, while $m$ and $s$ represent the margin and scale hyperparameters, respectively. With conventional FR training methods, only prototypes for real individuals, which can be denoted as $W_{R} = [w_{r}^1, w_{r}^2, ..., w_{r}^n]$, are necessary for training. In contrast, we include additional $k$ prototypes for virtual IDs, denoted as $W_{V} = [w_{v}^1, w_{v}^2, ..., w_{v}^k]$, which are used to generate facial images of nonexistent individuals in the real world. As a result, the prototype is defined as a linear transformation matrix $W \in \mathbb{R}^{(n+k) \times D}$, where $D$ refers to the dimension of the embedding features.

\begin{figure}[t]
\center
\begin{subfigure}{0.48\linewidth}
\includegraphics[width=\linewidth]{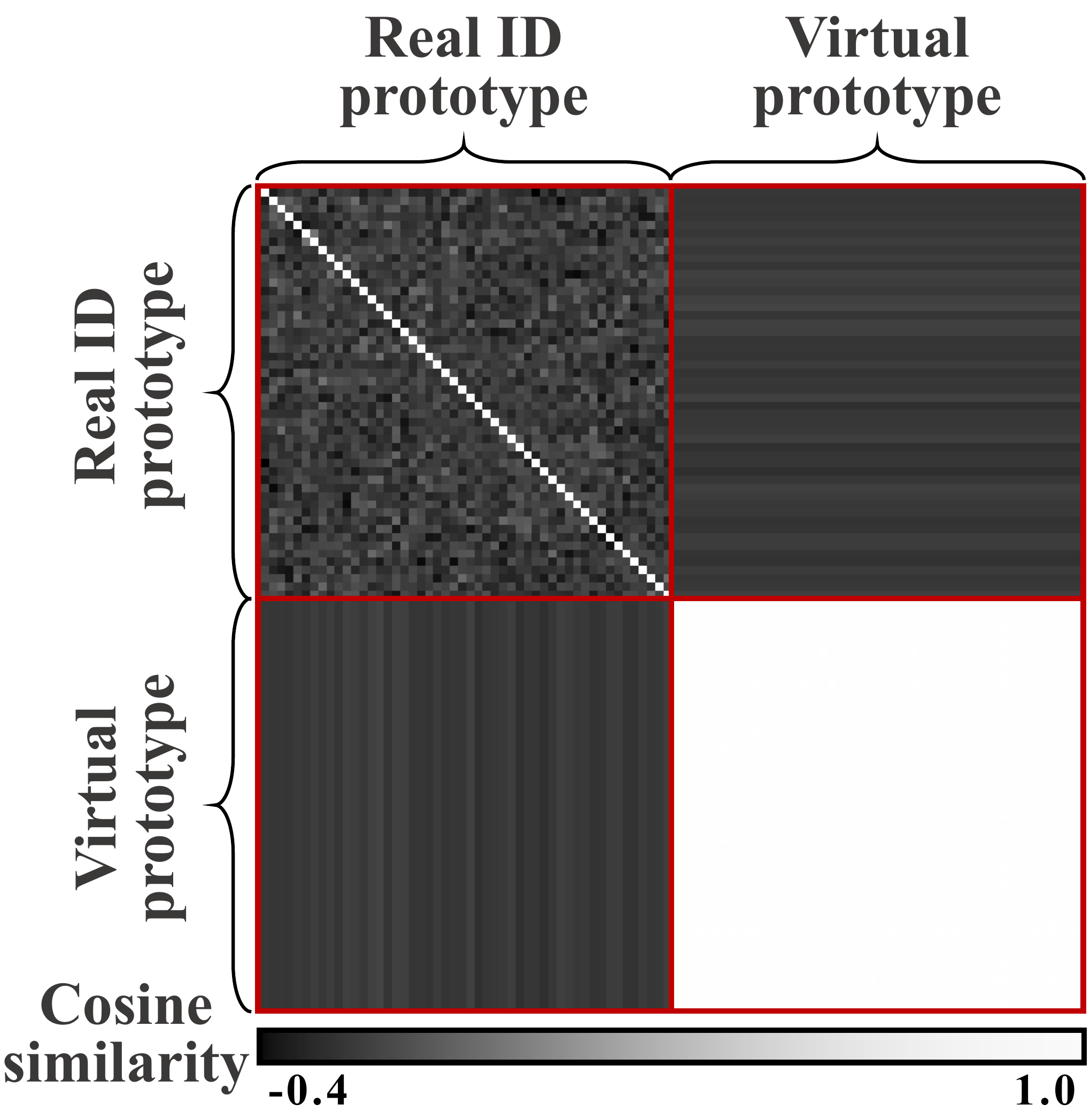}
\caption{Prototypes that are trained without virtual embeddings.}
\label{correlation1}
\end{subfigure}
\hfill
\begin{subfigure}{0.48\linewidth}
\includegraphics[width=\linewidth]{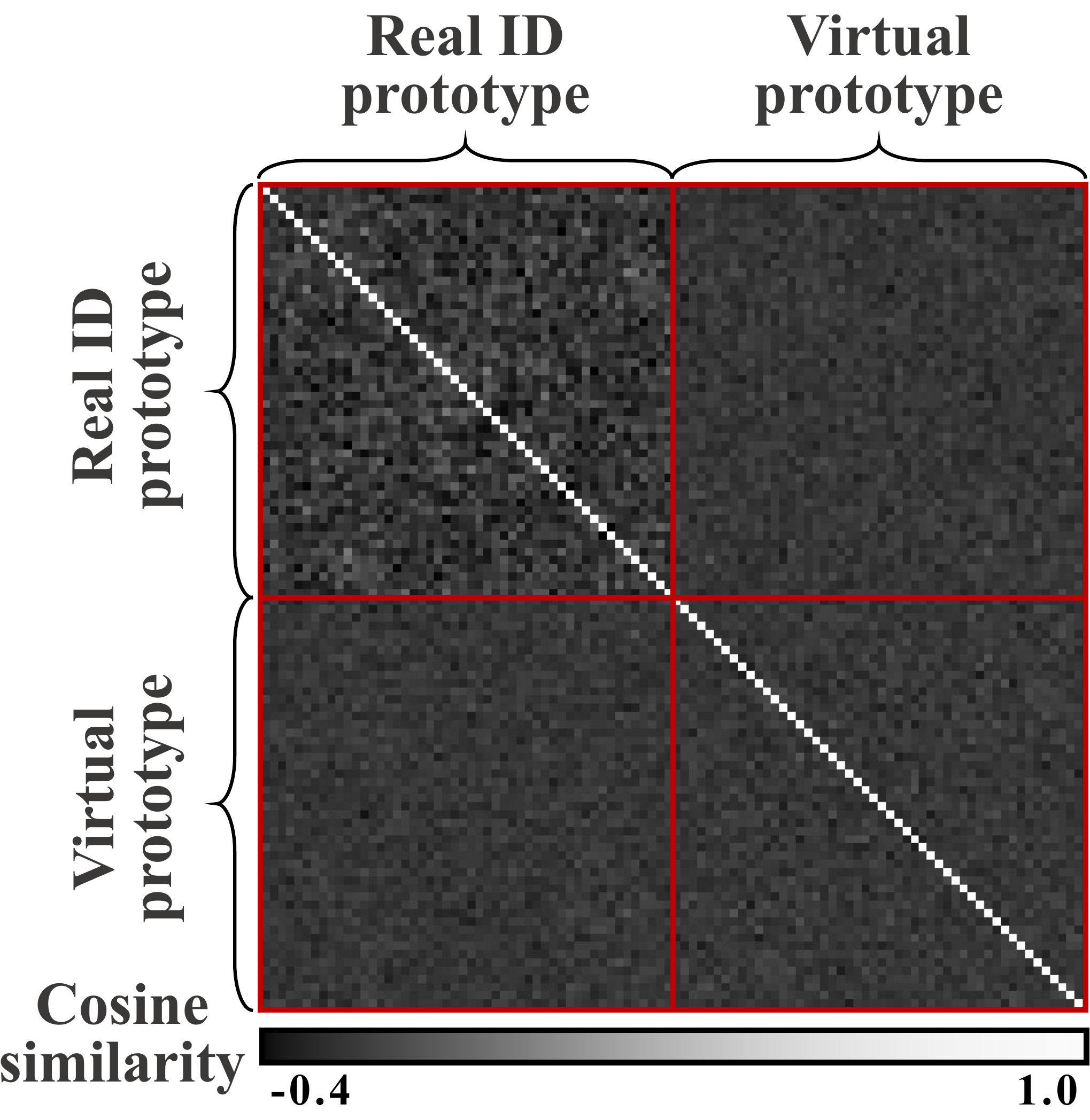}
\caption{Prototypes that are trained with both real and virtual embeddings.}
\label{Correlation2}
\end{subfigure}
\caption{
Changes in the similarity matrix of the prototypes from our method. Similarity values were min-max normalized.
}
\label{effect_virtual_sim_mat}
\end{figure}

However, due to the absence of a face image for virtual IDs, their prototypes cannot be updated to maximize the distance from each other with $L_{arc}$. Consequently, all virtual prototypes converge to a single point in the feature space when trained with the original $L_{arc}$ as shown in \cref{correlation1}.
To address this problem, we propose to use virtual feature embedding $f'_{\mathrm{FR}}(x_{j})$ to update the virtual prototypes $w_{v}^{j}$. The virtual embedding $f'_{\mathrm{FR}}(x_{j})$ corresponding to the virtual person ID: $j$ was generated as follows:
\begin{equation}
f'_{\mathrm{FR}}(x_{j}) = w_{v}^{j} + \mathcal{N}(0,\,1) \cdot \sigma,
\label{virtual_embedding}
\end{equation}
\begin{equation}
\sigma^2 =\frac{1}{b}\sum_{i=1}^{b}(f_{\mathrm{FR}}(x_{i})-w_{r}^i)^{2}, 
\label{virtual_embedding_sigma}
\end{equation}
where $b$ refers the mini-batch size. As can be seen from the equations, the virtual embedding $f'_{\mathrm{FR}}(x_{j})$ follows a distribution in which the standard deviation matches that of the real embeddings. The aggregate loss $L_{\mathrm{arc}}$ is calculated simultaneously using both the real embedding $f_{\mathrm{FR}}(x)$ and the virtual embedding $f'_{\mathrm{FR}}(x)$, allowing the virtual prototype $w_{v}^{j}$ to maintain minimal similarities with other prototypes, as illustrated in \cref{effect_virtual_sim_mat}. Since batch configuration affects the calculated standard deviation, we utilize the exponential moving average (EMA) to reduce this influence. The corrected standard deviation $\sigma$ for the current iteration $t$ is calculated as follows:
\begin{equation}
\sigma = \sigma_{(t)} \cdot \alpha + \sigma_{(t-1)} \cdot (1-\alpha).
\end{equation}
The hyperparameter $\alpha$ is set to 0.9. 
The number of virtual embedding, $b_{v}$ is determined based on the mini-batch size $b_{r}$, the number of virtual ID prototypes $k$, and the number of real ID prototypes $n$. In our study, we set $b_{v} = (k\times b_{r})/n$ so that both real and virtual prototypes can be updated evenly.

The overall pipeline of this stage is illustrated in \cref{fig:pipeline}. Note that, as can be seen in the figure, gradients for virtual embeddings have no effect on the backbone. We confirm that the similarity distribution between the virtual prototypes closely resembles the similarity distribution between the real prototypes, as shown in \cref{Correlation2}.

\vspace{5mm}
\subsection{Stage 2: Face Generation with Diffusion Model}

\begin{figure*}[t]
    \center
    \includegraphics[width=1.0\linewidth]{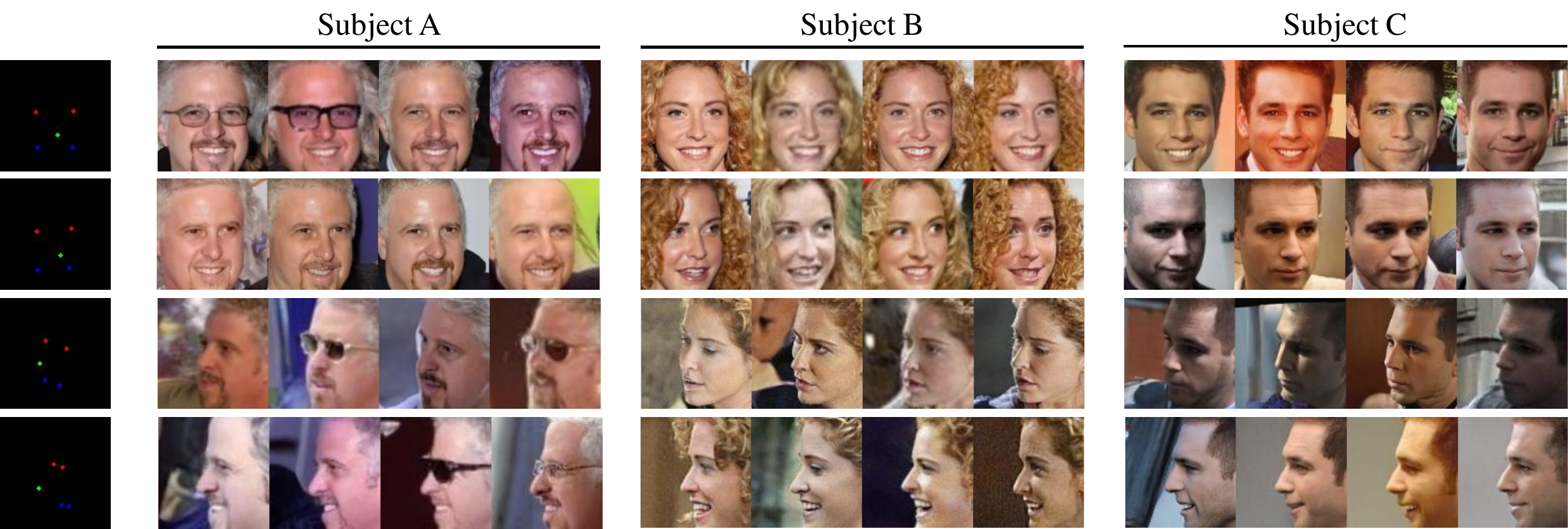}
    \caption{Virtual face generated in VIGFace (B). Each row lists facial images in different pose environments created using five facial landmarks. Our method can generate various conditions of face images, such as illumination, occlusion by accessories, and facial expressions, while controlling the pose variations of the face images.}
    \label{concat_pose_fig}
\end{figure*}

The next step is to synthesize face images using the diffusion model. 
To obtain the training dataset for the diffusion model, we utilize the FR model, which involves collecting pairs of images $x_{0}$ and their corresponding prototypes $w_{r}$. 
The input to our diffusion model includes the timestep $t$, the FR prototype vector $w_{r}$, the five facial landmark image $y$, and the noisy image $x_{t}$. 
In line with the approach proposed in the previous method~\cite{lin2024common}, our model predicts velocity~$v_{t}$ rather than noise~$\epsilon$ injected into $x_{t}$.
We adopt the DiT architecture~\cite{peebles2023scalable} using the design approach of the visual transformer (ViT)~\cite{dosovitskiy2010image}.
We modify the DiT model by incorporating the five facial landmarks~(including the left eye, the right eye, the nose, the left mouth corner, and the right mouth corner)~\cite{zhang2016joint} image. The five facial landmark images are acquired using RetinaFace~\cite{deng2019retinaface}, and employed as conditions to account for pose variations.
\cref{concat_pose_fig} shows the synthesized face images conditioned by five facial landmarks. The first column lists input landmark conditions in different pose environments. As illustrated in the figure, the proposed model is capable of producing a wide range of styles (like hair, glasses, and lighting) while also delivering various pose variations, all with a high degree of consistency.
In order to achieve both the generation of facial images and the synchronization of their feature space on the FR model, our diffusion model incorporates a constraint which aims to minimize the feature distance between the original image and the input prototypes as follows:
\begin{equation}
\min_{\theta}\mathbb{E}_{\epsilon, t}\| f_{\mathrm{FR}}(\hat{x}_{\theta}(x_{t}, t, w_r, y))) - w_r \|^2 _2 .
\label{FR_loss}
\end{equation}
We adopt classifier-free guidance~\cite{ho2021classifier} by randomly assigning zero values to condition embeddings, $w_r$, 10\% of time. Sampling is performed as follows:
\begin{multline}
\tilde{x}_{\theta}(x_{t}, t, W, y) = g \cdot x_{\theta}(x_{t}, t, W, y) \\ +  (1-g) \cdot x_{\theta}(x_{t}, t, y),
\label{classifier_free}
\end{multline}
where $x_{\theta}(x_{t}, t, w_r, y)$ and $x_{\theta}(x_{t}, t, y)$ are conditional and unconditional $x_0$-prediction, respectively and $g$ is \textit{guidance weight}.

\section{Experiments}
Our implementation details for the full framework are described in ~\cref{sec:implementation details}. We analyze our framework from two perspectives. Firstly, by training the FR model exclusively with facial images generated for virtual IDs, we demonstrated the capability of our model to serve as a viable alternative to real face datasets, addressing concerns such as label noise or privacy. Secondly, we train the FR model using both real images and generated face images simultaneously, showcasing the potential of our model as a data augmentation framework for face recognition.

\subsection{Virtual Identity Generation}

Since diffusion models focus on optimizing the visual aspect with input conditions, they tend to produce images that are high in consistency but lack of variance. One simple way to achieve a balance between consistency and variance is by adjusting the classifier-free guidance weight scale. Higher guidance enforces stronger conditioning of input labels. In other words, it results in more constant images but slightly similar samples. We observe that the proposed model performs best as the scale becomes w = 4.0.

We construct a toy example visualizing the feature space consists of three real persons and five virtual persons to demonstrate the efficacy of the synthesized images in \cref{tsne}. As seen in the figure on the left, the virtual embedding optimization obviously provides separable virtual prototypes from real individual clusters. The figure on the right illustrates that our synthesized images enforce high variance to the dataset. Synthesized images of real individuals filling the gap in the cluster provide intra-class variance to the subjects. Additionally, the cluster of generated images from virtual prototypes, while staying separate from other subjects, demonstrates the effectiveness of our framework for generating unique virtual human faces in high consistency.

\begin{figure}[t]
    \center
    \includegraphics[width=1.0\linewidth]{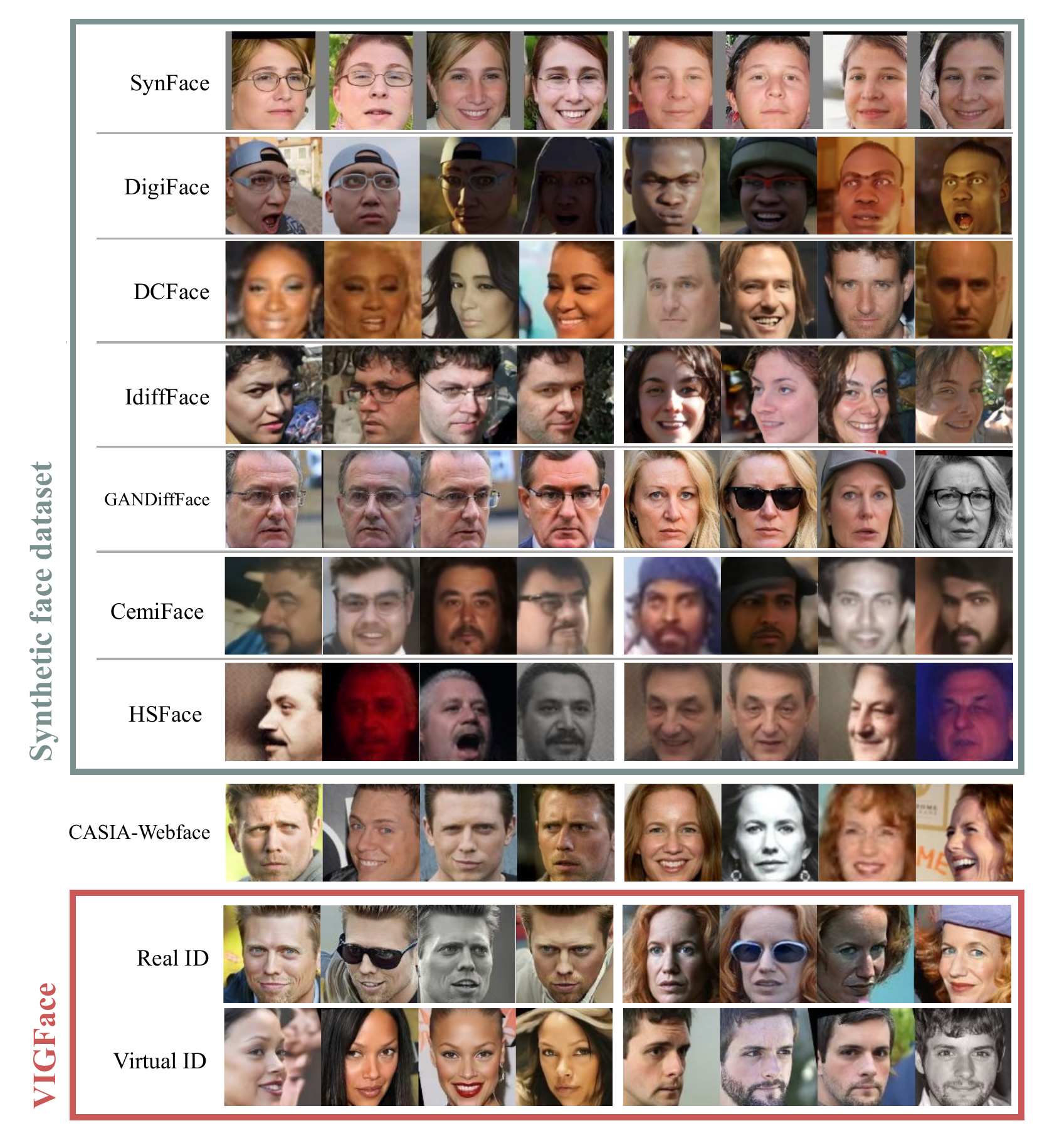}
    \caption{Comparison of the generated virtual ID images with the conventional methods~\cite{qiu2021synface, bae2023digiface, kim2023dcface, boutros2023idiff, melzi2023gandiffface,sun2025cemiface,wu2024vec2face} and our methods that are trained on CASIA-WebFace. For each synthetic dataset, we present two subjects in a single row.}
    \label{fig_results_compare}
\end{figure}

\cref{fig_results_compare} shows the qualitative comparison of conventional synthetic datasets and the proposed VIGFace. As shown in the figure, the facial images generated using the VIGFace model show remarkable uniformity in generating consistent virtual individuals, while also incorporating variations such as hair styles, accessories, makeup, and expressions. It is important to emphasize that the key characteristic of a training dataset to achieve a high-performance FR model is not the high quality of the images themselves, but rather the high consistency and variance of the generated facial images belonging to the same person.

\paragraph{Property of generated Dataset}

To compare the properties of the synthetic face datasets, we measured the 1) class consistency, 2) class separability, and 3) intra-class diversity of generated images. The class consistency reflects the uniformity of the samples in the same label condition. Consequently, the consistency of class $k$ ($\mathrm{C}_{k}$) was measured as follows:
\begin{equation}
C_{k} = \frac{1}{N^2}\sum_{j=1}^{N}\sum_{i=1}^{N}\frac{f(x_{i}) \cdot f(x_{j})}{\|f(x_{i}) \| \| f(x_{j}) \|}
\label{intra-class consistency},
\end{equation}
where $N$ is the number of images in a class $k$. Higher class consistency means that the samples are more uniform under the same label.

We also measured the class separability to assess the integrity of the dataset; in other words, to ensure that all subjects in the dataset are unique. The class separability for a class $k$ ($\mathrm{S}_{k}$) is measured as the average distance between the center of class $k$ and the centers of the negative classes, as follows:
\begin{equation}
\mathrm{S}_{k} = \frac{1}{K-1}\sum_{i=1, i \neq k}^{K}  1 - \frac{\overbar{f_{k}} \cdot \overbar{f_{i}}}{\|\overbar{f_{k}} \| \| \overbar{f_{i}} \|},
\label{inter-class consistency}
\end{equation}
where $K$ is the number of classes. $\overbar{f_{k}}$ represents the class center obtained by averaging the embedding vectors of the images that belong to class $k$. As shown in the formula, a high $\mathrm{S}_{k}$ indicates that the images are distinct from the negative classes.

The intra-class diversity measures how various the conditions of samples are under the same label. 
In particular, we focus on challenging scenarios that directly impact FR performance, such as pose variations, occlusions, or lighting conditions, rather than image style. Higher diversity indicates that the dataset covers a broad spectrum of cases, from easy to hard.
Motivated by this observation, we calculated the intra-class diversity of a class $k$ ($\mathrm{D}_{k}$) using the variance of Face Image Quality Assessment (FIQA) scores as follows:
\begin{equation}
\overbar{\operatorname{FIQA}_k} = \frac{1}{N}\sum_{i=1}^{N} \operatorname{FIQA}(x_i), 
\label{intra-class variance ser}
\end{equation}
%
%
\begin{equation}
\mathrm{D}_{k} = \frac{1}{N}\sum_{i=1}^{N}(\operatorname{FIQA}(x_i)-\overbar{\operatorname{FIQA}_k})^{2}, 
\label{intra-class variance}
\end{equation}
where $\operatorname{FIQA}(x_i)$ indicates the normalized CR-FIQA~\cite{boutros2023cr} score of the generated image $x_i$.

\begin{figure}[t]
\center
\begin{subfigure}[t]{0.51\linewidth}
\includegraphics[width=\linewidth]{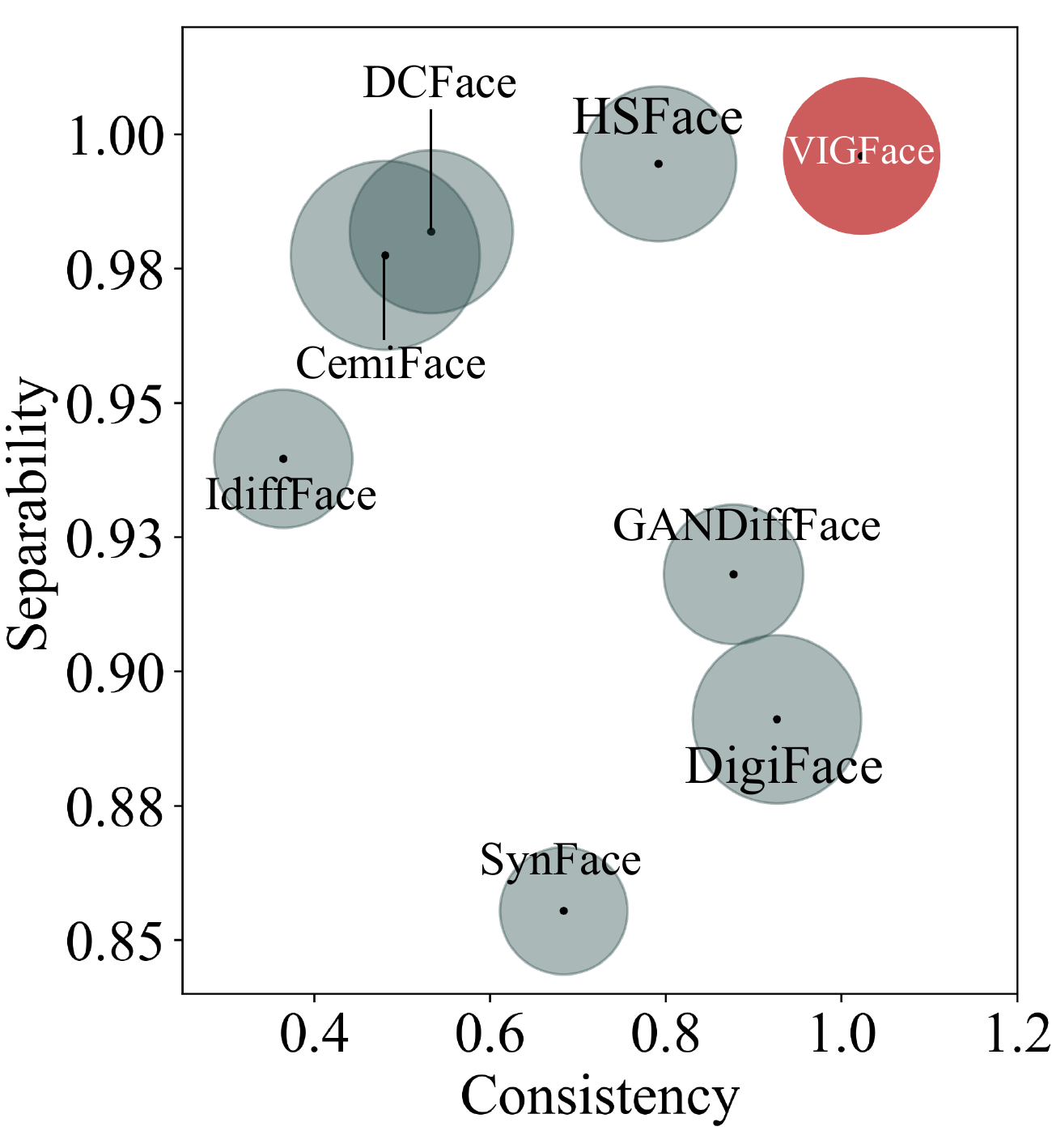}
\caption{The normalized properties of various synthetic datasets. The sizes of the circles indicate the intra-class diversity.}
\label{consist_div_fig}
\end{subfigure}
\hfill
\begin{subfigure}[t]{0.465\linewidth}
\includegraphics[width=\linewidth]{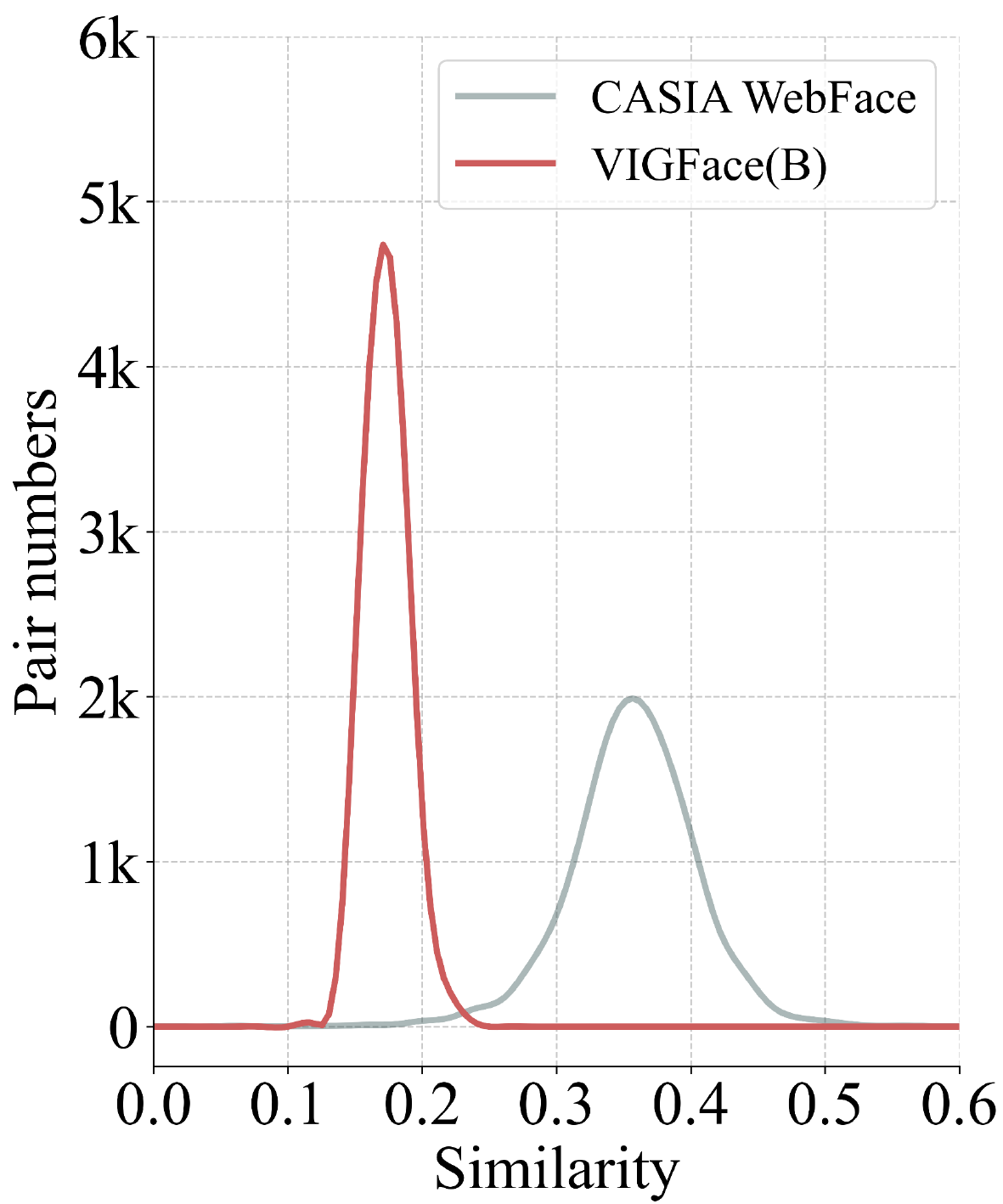}
\caption{Cosine similarity with the most similar negative class center in the real dataset (CASIA-WebFace).}
\label{fig_originality}
\end{subfigure}
\caption{Properties of VIGFace compared with those of synthesized datasets generated by previous methods.}
\label{fig_properties}
\end{figure}

\begin{table*}[t]
\centering
\resizebox{\linewidth}{!}{
\begin{tabular}{l|l|l|ccccc|c} 
\hline
\multirow{2}{*}{Method} & \multirow{2}{*}{Training Dataset} & \multirow{2}{*}{\begin{tabular}[c]{@{}l@{}}\# of Images\\(classes $\times$ variations)\end{tabular}} & \multicolumn{6}{c}{Verification Benchmarks} \\ 
\cline{4-9}
 &  &  & \multicolumn{1}{c|}{LFW} & \multicolumn{1}{c|}{CFP-FP} & \multicolumn{1}{c|}{CPLFW} & \multicolumn{1}{c|}{AgeDB} & CALFW & Avg. \\ 
\hline\hline
CASIA-Webface (Real) & - & 0.49M ($\approx$10.5K$\times47$) & 99.40 & 96.63 & 90.23 & 94.68 & 93.70 & 94.93 \\ 
\hline
SynFace~ & FFHQ & 0.5M (10K $\times$ 50) & 91.93 & 75.03 & 70.43 & 61.63 & 74.73 & 74.75 \\
DigiFace~ & 3D modeling & 0.5M (10K $\times$ 50) & 95.40 & 87.40 & 78.87 & 76.97 & 78.62 & 83.45 \\
DCFace~ & FFHQ+CASIA & 0.5M (10K $\times$ 50) & 98.55 & 85.33 & 82.62 & 89.70 & 91.60 & 89.56 \\
IDiffFace~ & FFHQ & 0.5M (10K $\times$ 50) & 98.00 & 85.47 & 80.45 & 86.43 & 90.65 & 88.20 \\
GANDiffFace~ & FFHQ & 0.5M (10K $\times$ 50) & 90.77 & 73.27 & 72.32 & 66.35 & 74.68 & 75.48 \\
ID$^3$~ & CASIA & 0.5M (10K $\times$ 50) & 97.68 & 86.84 & 82.77 & 91.00 & 90.73 & 89.80 \\
CemiFace~ & CASIA+WebFace4M & 0.5M (10K $\times$ 50) & \textbf{99.03} & 91.06 & \uline{87.62} & \uline{91.33} & 92.42 & \uline{92.30} \\
Arc2Face~ & WebFace42M & 0.5M (10K $\times$ 50) & 98.81 & \uline{91.87} & 85.16 & 90.18 & \uline{92.63} & 91.73 \\
HyperFace~ & WebFace42M & 0.5M (10K $\times$ 50) & 98.50 & 88.83 & 84.23 & 86.53 & 89.40 & 89.50 \\
HSFace10K~ & WebFace4M & 0.5M (10K $\times$ 50) & 98.87 & 88.97 & 85.47 & \textbf{93.12} & \textbf{93.57} & 92.00 \\
\rowcolor[rgb]{0.933,0.933,0.933} VIGFace(S), Ours & CASIA & 0.5M (10K $\times$ 50) & \uline{99.02} & \textbf{95.09} & \textbf{87.72} & 90.95 & 90.00 & \textbf{92.56} \\ 
\hline
DCFace~ & FFHQ+CASIA & 1.2M (20K$\times$ 50 + 40K$\times$ 5) & 98.58 & 88.61 & 85.07 & 90.97 & 92.82 & 91.21 \\
CemiFace~ & CASIA+WebFace4M & 1.2M (20K$\times$ 50 + 40K$\times$ 5) & 99.22 & 92.84 & 88.86 & 92.13 & 93.03 & 93.22 \\
Arc2Face~ & WebFace42M & 1.2M (20K$\times$ 50 + 40K$\times$ 5) & 98.92 & 94.58 & 86.45 & 92.45 & 93.33 & 93.14 \\
HSFace20K~ & WebFace4M & 1.0M (20K $\times$ 50) & 98.87 & 89.87 & 86.13 & \uline{93.85} & \uline{93.65} & 92.47 \\
HSFace300K~ & WebFace4M & 15M (300K $\times$ 50) & 99.30 & 91.54 & 87.70 & \textbf{94.45} & \textbf{94.58} & 93.52 \\
\rowcolor[rgb]{0.933,0.933,0.933} VIGFace(B), Ours & CASIA & 1.2M (60K$\times$ 20) & \textbf{99.48} & \uline{97.07} & \uline{90.15} & 93.62 & 92.88 & \uline{94.64} \\
\rowcolor[rgb]{0.933,0.933,0.933} VIGFace(L), Ours & CASIA & 6.0M (120K$\times$ 50) & \uline{99.33} & \textbf{97.31} & \textbf{91.12} & 93.82 & 92.95 & \textbf{94.91} \\
\hline
\end{tabular}
}
\caption{FR benchmark results trained with various virtual face datasets. All results except for CASIA-WebFace and VIGFace are obtained from the original paper. Our method is specified by the number of identities as small~(S), base~(B) and large~(L). FR backbone is IR-SE50 + AdaFace~\cite{kim2022adaface}. \textbf{Bold} and \underline{underline} indicates the best and the second best, respectively.} 
\label{bench_table1}
\end{table*}

The scores $\mathrm{C}_{k}$ and $\mathrm{S}_{k}$ are derived from a pre-trained ArcFace model that was trained using the Glint-360K dataset. \cref{consist_div_fig} shows the average values of the properties for all classes, normalized by the average values achieved with CASIA-WebFace. VIGFace achieves remarkable scores in aspects of class consistency and class separability compared to other methods.
SynFace exhibits the lowest separability due to its mix-up generation method. This fact indicates that SynFace can generate only a limited number of subjects as mentioned in the introduction. DigiFace achieves strong consistency, due to its 3D rendering methods, but its image style may differ significantly from real datasets, leading to lower accuracy in real-world applications. GANDiffFace demonstrates inadequate FIQA diversity, indicating limited variations in factors such as pose, lighting, and occlusion.
DCFace, which uses a diffusion model, shows separability similar to that of VIGFace but with lower consistency. 

In comparisons of SOTA models, HSFace, despite using a more refined dataset, comparatively lacks consistency. CemiFace, trained on the same dataset as VIGFace, shows low consistency and separability. Both methods also exhibit inferior qualitative uniformity within the same class, as illustrated in \cref{fig_results_compare}.
Our approach uses prototypes instead of FR embedding vectors to disentangle identity features from other characteristics, allowing VIGFace to generate subjects with greater consistency by focusing on identity features. Moreover, it enhances separability by leveraging the pre-computation of feature orthogonality. We included further analysis on the properties of the datasets and provided examples in~\cref{sec:suppl_failure case}.

\paragraph{Identity leakage from real human}
To claim complete privacy-free, it is necessary to prove that no real IDs or training images are included in the synthesized dataset. In this reason, we demonstrate that the generated face images represent non-existent humans by querying the most similar face in the CASIA-WebFace dataset. For comparison, we also present the similarity values of the nearest negative class in CASIA-WebFace itself. As shown in \cref{fig_originality}, VIGFace(B)'s top-1 similarity score for a real human is lower than the nearest negative class similarity in CASIA-WebFace. This supports the claim that each synthesized image depicts a non-existent human face and there is no identity leakage from the training data. 
In contrast, we observed that the conventional SOTA methods show identity leakage. As CemiFace~\cite{sun2025cemiface} samples identity embeddings from the WebFace4M dataset, it generates individuals that exhibit resemblance to those in WebFace4M. Vec2Face~\cite{wu2024vec2face} fails in creating novel identities, so that it produces almost same persons in WebFace4M. Furthur analysis and examples are reported in~\cref{app:id_leak}.
In this paper, we avoid thresholding or selective sampling, which could harm fair comparisons with previous methods.

\begin{table*}[t]
\centering
\resizebox{\linewidth}{!}{%
\begin{tabular}{c|c|c|c|ccccc|c|c} \hline
\multicolumn{4}{c|}{Condition} & \multicolumn{7}{c}{Verification Benchmarks} \\ \hline
\multirow{2}{*}{Method} & \multirow{2}{*}{Real Image} & \multicolumn{2}{c|}{Synthetic Image} & \multirow{2}{*}{LFW} & \multirow{2}{*}{CFP-FP} & \multirow{2}{*}{CPLFW} & \multirow{2}{*}{AgeDB} & \multirow{2}{*}{CALFW} & \multirow{2}{*}{Avg.} & \multirow{2}{*}{$\varDelta$} \\ \cline{3-4}
 &  & Real ID & Virtual ID &  &  &  &  &  &  &  \\ \hline\hline
CASIA-WebFace & \checkmark &  &  & 99.40 & 96.63 & 90.23 & 94.68 & 93.70 & 94.93 & - \\ \hline\hline
DigiFace & \checkmark &  & \checkmark & 99.37 & 97.51 & 90.92 & 94.95 & 93.77 & 95.30~ & +0.37 \\
DCFace (1.2M) & \checkmark &  & \checkmark & 99.43 & 96.97 & 90.33 & 95.20 & 94.38 & 95.26~ & +0.33 \\
iDiffFace & \checkmark &  & \checkmark & 99.58 & 97.04 & 90.40 & 94.78 & 94.00 & 95.16~ & +0.23 \\
GANDiffFace & \checkmark &  & \checkmark & 99.52 & 96.61 & 90.30 & 93.98 & 93.57 & 94.80~ & -0.13 \\
HSFace20K & \checkmark &  & \checkmark & 99.60 & 97.41 & 91.07 & 95.48 & \textbf{94.40} & 95.59 & +0.66 \\ \hline
\multirow{3}{*}{VIGFace(B)} & \checkmark & \checkmark &  & 99.45 & 97.23 & 90.78 & 95.25 & 93.22 & 95.19 & +0.26 \\ 
 & \checkmark &  & \checkmark & 99.55 & 98.03 & \textbf{91.80} & 95.73 & 94.12 & 95.85 & +0.92 \\
 & \checkmark & \checkmark & \checkmark & \textbf{99.70} & \textbf{98.10} & 91.57 & \textbf{95.85} & 94.38 & \textbf{95.92} & \textbf{+0.99} \\ \hline
\end{tabular}
}
\caption{Augmented FR performance results for various condition of synthetic dataset. The accuracies (\%) for LFW, CFP-FP, CPLFW, AgeDB-30, CALFW, and the average benchmark accuracies are reported. FR backbone is IR-SE50 + AdaFace~\cite{kim2022adaface}.}
\label{bench_table_ablation}
\end{table*}

\subsection{Evaluation}
\label{eval_FR}
In this section, we train the FR model with VIGFace to compare it to conventional methods. Detailed hyperparameters for training the FR network can be found in~\cref{suppl_hyperparameter_FR_training}.
\paragraph{VIGFace as Virtual Dataset}
We compare the performance of the FR model trained with generated facial images for virtual IDs using VIGFace with conventional methods. For the experiment, we set the number of virtual IDs to $10K$, $60K$ and $120K$. To ensure a fair comparison, we established image counts of 0.5M, 1.2M, and 6.0M, which reflects the scale of the real CASIA-WebFace dataset and previous synthetic approaches, respectively. In \cref{bench_table1}, we present the 1:1 verification accuracy~(\%) on five benchmarks~\cite{huang2008labeled, sengupta2016frontal, zheng2018cross, moschoglou2017agedb, zheng2017cross}. 

As shown in the table, VIGFace outperforms conventional methods in the average accuracy of five verification benchmarks. In particular, VIGFace achieves equal or even better performance on CFP-FP and CPLFW compared to the model trained with the real dataset, CASIA-WebFace. This indicates that our method benefits from its pose-variable generation capabilities. 
CemiFace shows notable performance on the AgeDB and CALFW benchmarks, but they sampled identity embeddings from the WebFace4M~\cite{zhu2021webface260m} dataset. Consequently, the individuals included in the CemiFace dataset may still have privacy concerns.
HSFace10K also delivers outstanding results among conventional methods. However, it is important to note that HSFace benefits from training on the extensive WebFace4M dataset, which boasts twice as many images and quadruples the number of identities when compared to CASIA-WebFace. VIGFace(L) outperforms HSFace300K even with a smaller number of images.
As a result, when the FR model was trained using VIGFace, it achieved performance comparable to a model trained using a real dataset in terms of average accuracy without any external identity sampling. This suggests that our virtual dataset can serve as a full replacement for the CASIA-WebFace dataset to train the FR model, while avoiding privacy concerns.

\paragraph{VIGFace as Data Augmentation}
To demonstrate the efficacy of VIGFace as an augmentation framework, we evaluate the accuracy of the FR model trained on real and synthetic images. We utilized the dataset that was uploaded to the official repository. \cref{bench_table_ablation} shows the performance change of the trained FR model using various combinations of datasets. DigiFace, which utilizes unique 3D modeling technology, has demonstrated improved benchmark performance and offers advantages from a data-augmentation perspective. However, DigiFace struggles to perform well on its own, devoid of real data~\cite{bae2023digiface}. This is due to their failure to precisely replicate the appearance of real-world facial images, making it impractical to present a privacy-free synthetic dataset. HSFace shows a notable improvement among conventional methods. However, it utilized WebFace4M for training, which may make the comparison with others unfair. The boosted accuracy using VIGFace, which outperforms conventional methods, demonstrates that VIGFace can accurately mirror real facial data and achieve synergy by blending them. In particular, augmented images of real IDs improve the results of the CFP-FP and CPLFW benchmarks. This observation suggests that the use of five-point facial landmarks in the conditioning method can create a variety of posed facial images, substantially improving the FR model's ability to understand pose variations.
Consequently, VIGFace not only effectively solves the privacy issue, but can also be used with real data as part of the augmentation process. When trained with the VIGFace dataset, the FR backbone achieves better generalization and higher performance, in contrast to conventional methods that achieved only marginal performance improvements.

\cref{fig_results_compare} illustrates the results that synthesize variations of real and virtual IDs. Since the training process is significantly biased not just by a lack of subject but also by inner-class variance, increasing the variance within the class is crucial. 
The results demonstrate that the proposed approach can generate a variety of conditions for the existing real ID without any help from external data. Since VIGFace can generate variance images of real individuals, we performed experiments on each type of augmentation, \ie synthetic images of real and virtual ID. In particular, we extend the long-tailed real-ID class, which contains fewer than 50 images, to 50 images. With this method, additional 0.15M images were added as real ID synthesis. As can be seen in \cref{bench_table_ablation}, the augmented dataset for real ID shows improved FR performance on most benchmarks. This indicates that our method can increase the intra-class diversity of dataset, a critical factor in achieving high FR performance. 
As a result, the FR model that utilizes the entire set of augmented data exhibits the best performance, benefiting from enhanced intra-class diversity and inter-class variance.

\section{Conclusions}
This paper presents an effective method for creating a synthetic face dataset that guarantees unique virtual identities. Synthesized facial data can serve as a solution for various challenges faced by traditional real datasets, such as high expenses, inaccuracies and biases in labeling, and concerns regarding privacy.
To this end, we propose the Virtual Identity Generation framework and demonstrate that it
can generate not only realistic but also diverse facial images of virtual individuals, significantly narrowing the performance gap with FR models trained on real data. Furthermore, the model exhibits superior performance when trained on a combination of VIGFace and existing real data compared to models trained solely on real data. This confirms that our proposed method has potential as an effective augmentation technique. We will publicly distribute the virtual face dataset created by VIGFace and believe that these new virtual data will contribute to resolving the privacy issues inherent in face recognition training dataset.

{
    \small
    \bibliographystyle{ieeenat_fullname}
    \bibliography{main}
}

\clearpage
\appendix
\setcounter{page}{1}

\maketitlesupplementary
%
\begin{table}
\centering
\resizebox{\linewidth}{!}{
\begin{tabular}{c|c} \hline
\multicolumn{2}{c}{Face Recognition Model Training Configurations} \\ \hline\hline
Head & AdaFace \\
Margin ($m$) & 0.4 \\
Scale ($s$) & 64 \\
Augmentation & Random Erase $\times$ Rescale $\times$ Jitter \\
Augmentation ratio & 0.2 $\times$ 0.2 $\times$ 0.2 \\
Reduce LR epochs (S, L) & [24, 30, 36], [12, 20, 24]  \\
Epochs (S, L) & 40, 26 \\
Backbone & IR-SE50 \\
Batch Size & 512 \\
Initial Learning Rate & 0.1 \\
Weight Decay & 5e-4 \\
Momentum & 0.9 \\
FP16 & True \\
Optimizer & SGD \\ \hline
\end{tabular}
}
\caption{Configurations for training the FR network. (S) and (L) represent training using the small dataset ($< 1.0$M) and the large dataset ($\geqq 1.0$M), respectively.}
\vspace{-5mm}
\label{fr_config_large}
\end{table}
\section{Implementation Details}\label{sec:implementation details}

\paragraph{For stage 1:} 
We use a modified ResNet-50~\cite{deng2019arcface} and ArcFace~\cite{deng2019arcface} to train the face recognition (FR) backbone. The CASIA-WebFace~\cite{huang2012learning} dataset serves as training data. Following the alignment method in~\cite{liu2017sphereface}, the facial images are aligned with a resolution of $112\times112$. After alignment, the pixel values are normalized, with both the mean and standard deviation set to 0.5. We set the mini-batch size for the real dataset to 512 and use Stochastic Gradient Descent (SGD) as the optimizer, with a weight decay of 5e-4 and momentum of 0.9. The initial learning rate is set to 0.1 and divided by 0.1 in the 24th, 30th, and 36th epochs, with training concluding at the 40th epoch. The ArcFace hyperparameters for the margin $m$ and the scale factor $s$ are set to 0.5 and 30, respectively.
%
\paragraph{For stage 2:}
To generate synthetic face images, we explore the best setting for the diffusion model. To facilitate comparison, we utilized the widely adopted DiT-B model for all experiments. Since the traditional FR model typically employs a resolution of $112\times112$ for face images, we set the window size of the DiT patch extractor to 4. We follow the publicly released implementation of DDIM~\cite{song2020denoising} with the cosine noise scheduler~\cite{chen2023importance}. Following the approach of the previous method~\cite{lin2024common}, we employ the enforced zero terminal SNR and trailing timesteps. The diffusion model is trained for 5M iterations with a batch size of 512 using AdamW Optimizer~\cite{kinga2015method, loshchilov2017decoupled} with a learning rate of 1e-4. For sampling, we used classifier-free guidance implementation~\cite{ho2021classifier} in 50 time steps.

%
\begin{figure*}[ht]
\center
\begin{subfigure}[t]{0.245\linewidth}
\includegraphics[width=\linewidth]{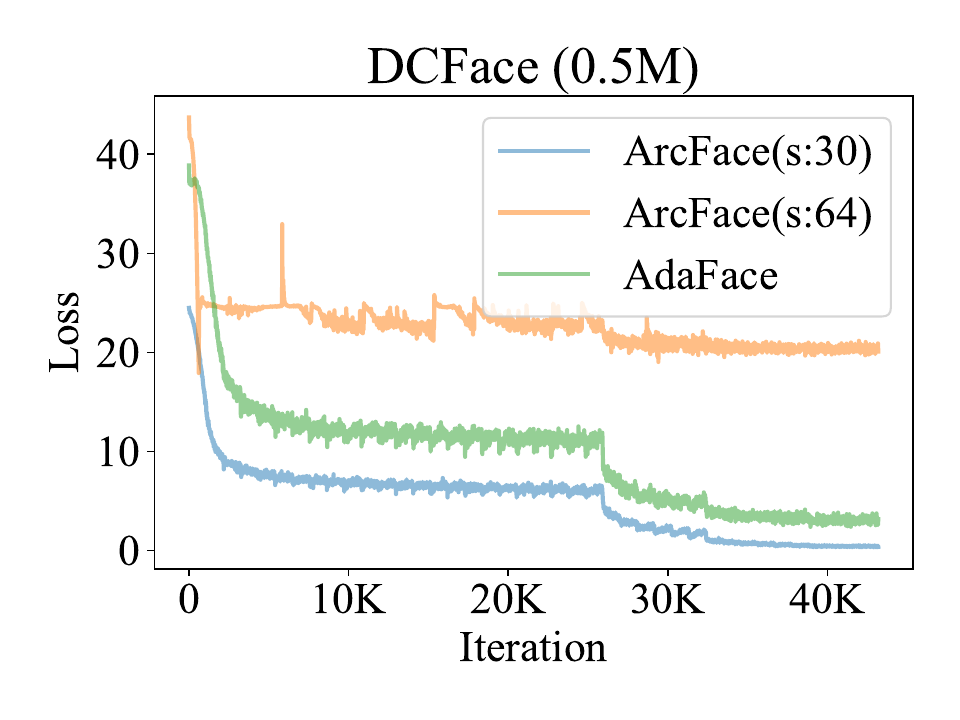}
\end{subfigure}
\begin{subfigure}[t]{0.245\linewidth}
\includegraphics[width=\linewidth]{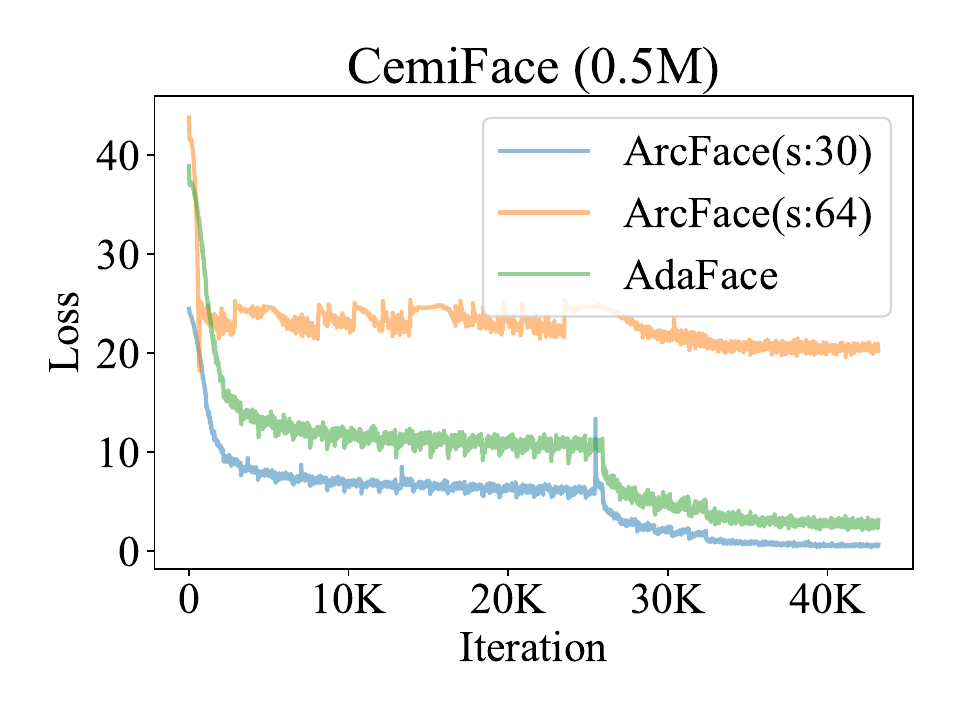}
\end{subfigure}
\begin{subfigure}[t]{0.245\linewidth}
\includegraphics[width=\linewidth]{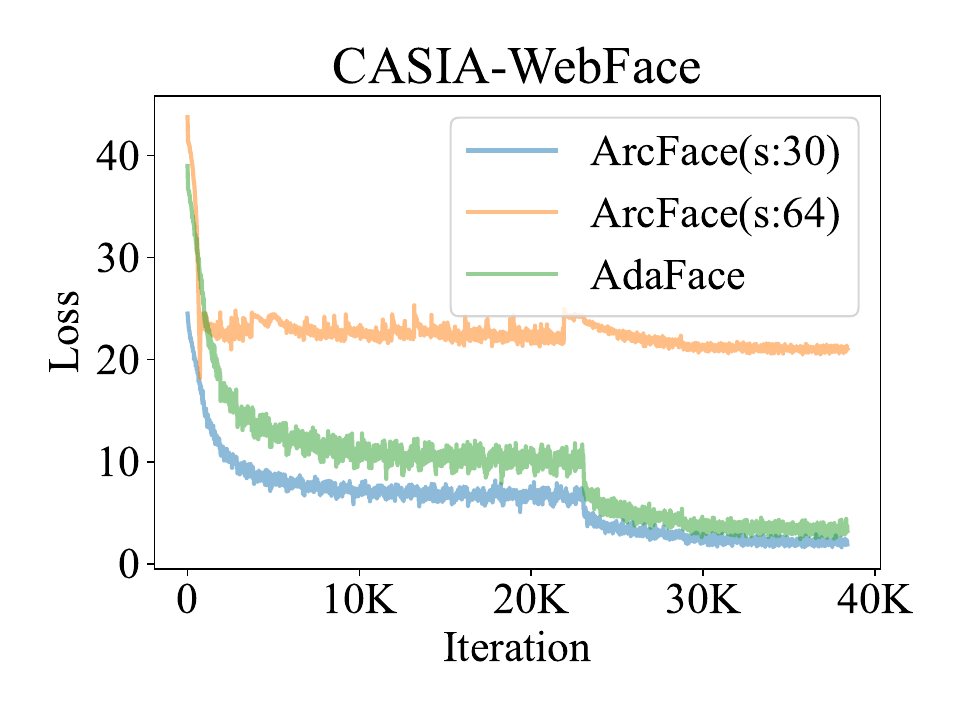}
\end{subfigure}
\begin{subfigure}[t]{0.245\linewidth}
\includegraphics[width=\linewidth]{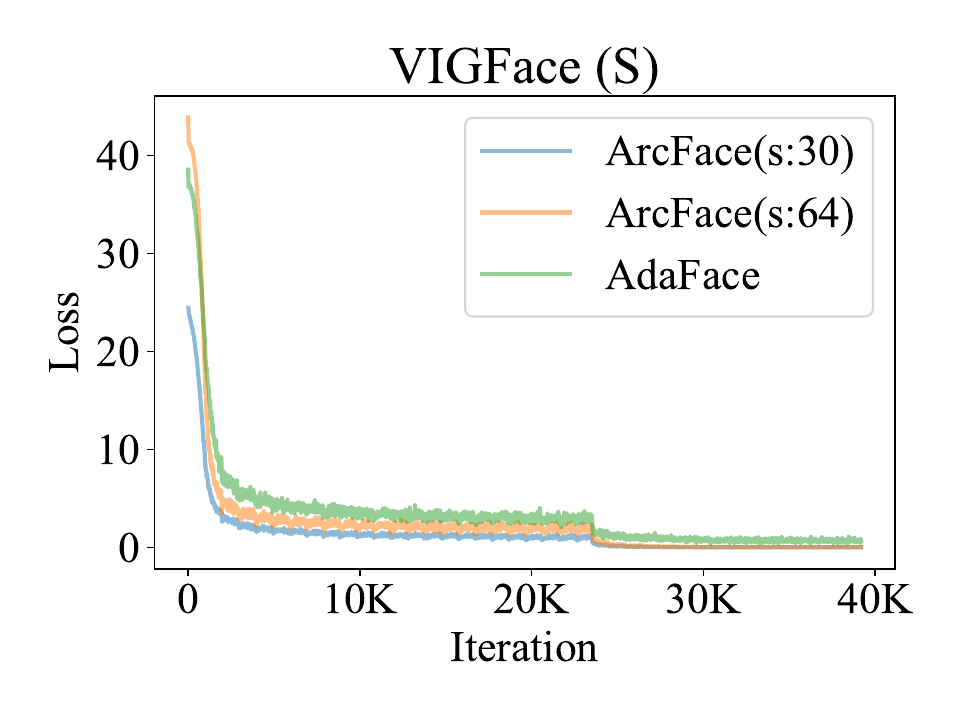}
\end{subfigure}
\caption{Loss log during training of FR backbone.}
\label{FR_backbone_log}
\end{figure*}

\paragraph{Implementation details for FR training:}
\label{suppl_hyperparameter_FR_training}
We provide the configurations for the FR network training that are used in ~\cref{eval_FR}. We strongly refer to the training methods proposed in~\cite{arc2face,kim2022adaface,wu2024vec2face} to set the hyperparameters. Detailed configurations can be found in~\cref{fr_config_large}. We adjusted training epochs and learning rate scheduling strategies based on the dataset size. Following~\cite{kim2022adaface}, we employ random erasing, rescaling, and color jittering as data augmentation, especially when training with AdaFace. Random erasing is applied by filling randomly selected regions with pixel values of 0. The erased region size is randomly set between 0.02 and 0.33 times the original image width, with an aspect ratio varying between 0.3 and 3.3. For rescaling, each face image is first shrunk and then restored to its original dimensions. The shrinking ratio is randomly selected between 0.2 and 1.0 times the original image width. To ensure diversity, we randomly apply one of the following interpolation methods during shrinking and restoration: nearest-neighbor interpolation, bilinear interpolation, bicubic interpolation, pixel area relation interpolation, and Lanczos4 interpolation. For color jittering, we randomly adjust the brightness, contrast, and saturation of the input image, with each factor modified within a range of 0 to 0.5 relative to the original image.
\section{FR training using unified configuration}
\label{sec:unified_configuration}
Training configuration, such as training loss, augmentations, and hyperparameters, is vital to the performance of FR models. However, conventional works have conducted the training based on their own configuration. In this section, we present reproduced experiments trained under the unified implementation except the training dataset in \cref{table_multi}. As the training code of conventional methods for SFR is not released, we utilize margin($m$) with 0.5, and scaling factor($s$) with 30 and 64, following the original paper~\cite{deng2019arcface} settings.

As shown in~\cref{FR_backbone_log}, we observed that ArcFace with a scaling factor ($s$) of 64 failed to train stably in the conventional datasets due to the gradient explosion. The high $s$ makes the softmax operation more steep near the decision boundary~\cite{kim2022adaface}. This leads to a gradient explosion during training when the data includes unrecognizable samples or label noise. CASIA-WebFace is known to include around $9.3\%-13.0\%$ of label noise data~\cite{wang2018devil}. In contrast, since VIGFace has high consistency, it is free from label noise and enables stable backbone training.

Note that the entire framework in VIGFace was trained using only CASIA-WebFace, without any external datasets or a pre-trained large model such as CLIP~\cite{radford2021learning}.

\begin{table}[t!]
\centering
\resizebox{1.0\linewidth}{!}{
\begin{tabular}{l|c|c|cll} \hline
\multirow{2}{*}{Training Dataset} & \multirow{2}{*}{Method} & \multirow{2}{*}{\begin{tabular}[c]{@{}c@{}}Average\\Accuracy\end{tabular}} & \multicolumn{3}{c}{IJB-C~\cite{maze2018iarpa}} TPR@FPR\\ \cline{4-6}
 &  &  & \multicolumn{1}{l}{1e-4} & 1e-3 & 1e-2 \\ \hline\hline
\multirow{3}{*}{CASIA-Webface} & Arc. (s:30) & 94.28 & 83.44 & 91.22 & 96.22 \\
 & Arc. (s:64) & 70.16 & 15.07 & 29.26 & 50.30 \\
 & Ada. & 94.93 & 77.65 & 92.66 & 96.89 \\ \hline\hline
\multirow{3}{*}{DCFace (0.5M)} & Arc. (s:30) & 84.80 & 60.29 & 76.42 & 88.97 \\
 & Arc. (s:64) & 71.50 & 11.37 & 22.62 & 41.07 \\
 & Ada. & 89.07 & 77.24 & 87.51 & 94.01 \\ \hline
\multirow{3}{*}{CemiFace (0.5M)} & Arc. (s:30) & 85.57 & 37.85 & 77.33 & 90.09 \\
 & Arc. (s:64) & 72.88 & 18.92 & 33.81 & 54.94 \\
 & Ada. & 90.54 & 83.11 & 90.41 & 95.39 \\ \hline
\multirow{3}{*}{HSFace10K} & Arc. (s:30) & 85.39 & 73.94 & 82.18 & 89.67 \\
 & Arc. (s:64) & 71.22 & 14.16 & 27.24 & 45.81 \\
 & Ada. & 90.22 & 86.10 & 91.04 & 94.92 \\ \hline
\rowcolor[rgb]{0.933,0.933,0.933} {\cellcolor[rgb]{0.933,0.933,0.933}} & \multicolumn{1}{l|}{Arc. (s:30)} & 91.19 & \multicolumn{1}{l}{69.27} & 82.64 & 91.78 \\
\rowcolor[rgb]{0.933,0.933,0.933} {\cellcolor[rgb]{0.933,0.933,0.933}} & Arc. (s:64) & 92.37 & 72.53 & 85.14 & 92.98 \\
\rowcolor[rgb]{0.933,0.933,0.933} \multirow{-3}{*}{{\cellcolor[rgb]{0.933,0.933,0.933}}VIGFace (S)} & Ada. & 92.56 & 80.00 & 89.22 & 94.99 \\ \hline\hline
\rowcolor[rgb]{0.933,0.933,0.933} {\cellcolor[rgb]{0.933,0.933,0.933}} & Arc. (s:30) & 94.00 & 80.33 & 88.87 & 94.69 \\
\rowcolor[rgb]{0.933,0.933,0.933} {\cellcolor[rgb]{0.933,0.933,0.933}} & Arc. (s:64)& 94.27 & 81.69 & 89.79 & 95.15 \\
\rowcolor[rgb]{0.933,0.933,0.933} \multirow{-3}{*}{{\cellcolor[rgb]{0.933,0.933,0.933}}VIGFace (B)} & Ada. & 94.64 & 83.29 & 91.22 & 95.97 \\ \hline
\end{tabular}
}
\caption{Re-implemented FR benchmark results trained under unified configurations. The FR models used in the table were trained with the same implementation details, following original papers~\cite{deng2019arcface, kim2022adaface}.}
    \label{table_multi}
\end{table}

\section{Stage 1 : Virtual prototype assignment}
\label{sec:stage1_analysis}

\paragraph{Performance of backbone in Stage 1: } 
We evaluate the performance of the trained backbone achieved at the end of stage 1. Given that our approaches employ virtual prototypes, it is essential to guarantee that the virtual prototype does not affect the FR backbone training. As can be seen in~\cref{stage1_performance}, our approach maintains the benchmark face recognition performance with negligible changes.

\paragraph{Pre-assigning vs Sampling:}
In Stage 1, VIGFace employs a pre-assignment strategy for virtual prototypes within the feature space, maximizing inter-prototype distances through the ArcFace loss. The ArcFace loss explicitly enforces a substantial margin between distinct identities in the feature space, ensuring that virtual prototypes are highly discriminative. This potentially yields superior identity separation compared to random sampling approaches, which rely on fixed similarity thresholds. Our theoretical analysis for the high separation of virtual identities is also supported by our property analysis in~\cref{consist_div_fig}. In addition, since the embedding vectors used in the generative model are closely aligned, VIGFace produces synthetic faces that follow the actual data distribution better than the random sampling method.

Although training virtual prototypes for a huge number of identities (\eg millions or billions) requires a large-scale classifier, making VIGFace computationally intensive compared to sampling methods. However, this challenge can be mitigated by integrating techniques such as Partial-FC~\cite{an2021partial}, which optimizes computational efficiency while maintaining performance.

\begin{table}[t]
\centering
\resizebox{\linewidth}{!}{
\begin{tabular}{l|c|c|c|c} \hline
Dataset & Real Prototype & Virtual Prototype & \begin{tabular}[c]{@{}c@{}}Average\\Accuracy\end{tabular} & IJB-C \\ \hline\hline
\multirow{2}{*}{CASIA-Webface} & \multirow{2}{*}{10.5K} & - & 94.28 & 83.44 \\
 &  & 60K & 94.30 & 83.46 \\ \hline\hline
\end{tabular}
}
\caption{Benchmark results of the stage 1: FR model which employs different number of virtual prototypes. For IJB-C, TPR@FPR=1e-4 is reported.}
\label{stage1_performance}
\end{table}

\paragraph{Scalability of virtual prototype:} Our method demonstrates that all virtual prototypes can nearly maintain orthogonality after stage 1. From probabilistic geometry~\cite{brauchart2018random}, if there are $N$ identities following a uniform distribution, the minimum cosine distance between two vectors in a feature space of dimension $d$ can be approximated as follows:

\begin{equation}
\cos(\theta) \approx \sqrt{\frac{\log N}{d}}.
\end{equation}
This implies that the 512 feature dimension, which is used in our experiments, is large enough to sort $10^8$ identity features while preserving nearly orthogonal.

\begin{figure*}[t]
\center
\begin{subfigure}[t]{0.4\linewidth}
\includegraphics[width=1.0\linewidth]{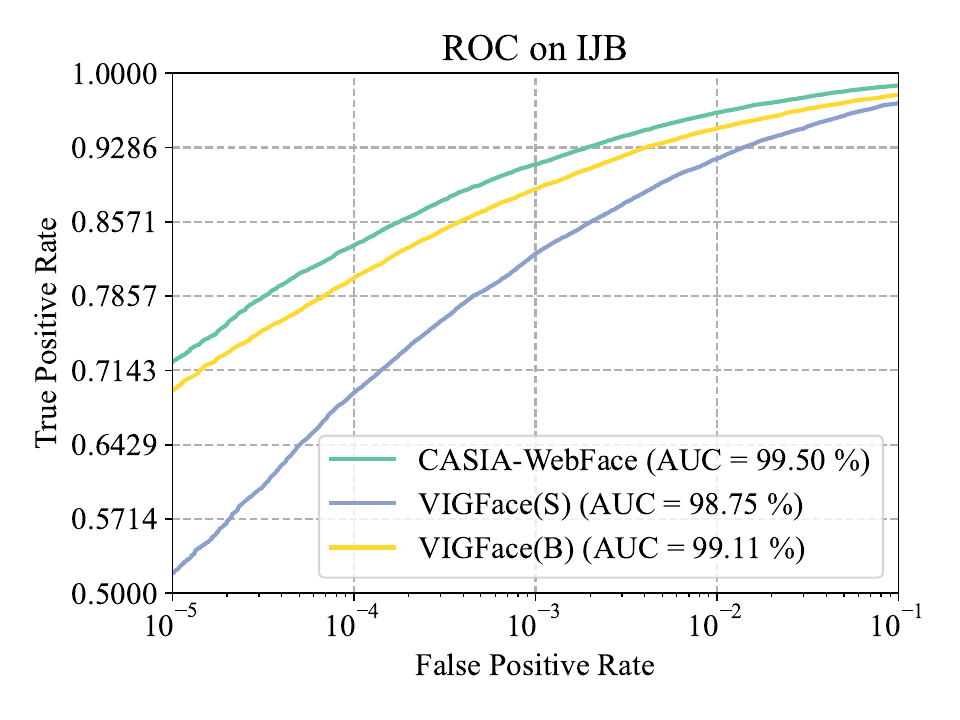}
\caption{IR-SE50 + ArcFace (s:30).}
\label{roc_ijbc}
\end{subfigure}
\begin{subfigure}[t]{0.4\linewidth}
\includegraphics[width=1.0\linewidth]{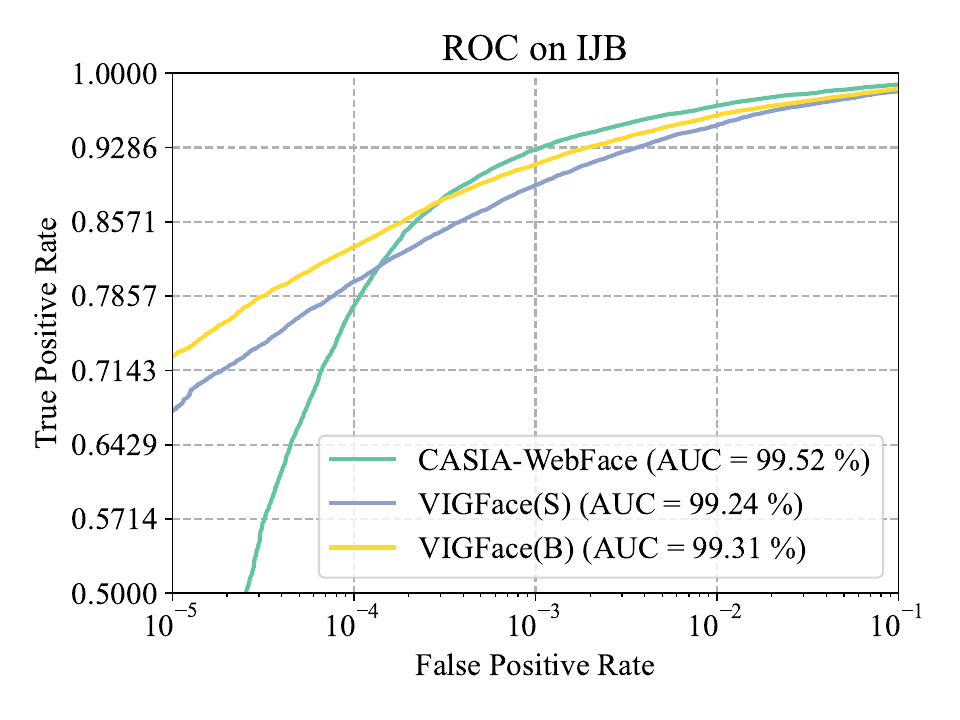}
\caption{IR-SE50 + AdaFace.}
\label{roc_ijbc}
\end{subfigure}
\caption{ROC on IJB-C benchmarks. }
\label{roc_ijb}
\end{figure*}

\section{ROC of IJB-C benchmarks}
We compare the Receiver Operating Characteristic (ROC) curves using the IJB-C~\cite{maze2018iarpa} dataset. \cref{roc_ijbc} presents the ROC curves of IJB-C for VIGFace. We observe that VIGFace outperforms real datasets at low FPR ($< 1e-4$) on AdaFace trained models.

\section{Face Image Quality distribution}
\begin{figure}[ht!]
\center
\includegraphics[width=0.85\linewidth]{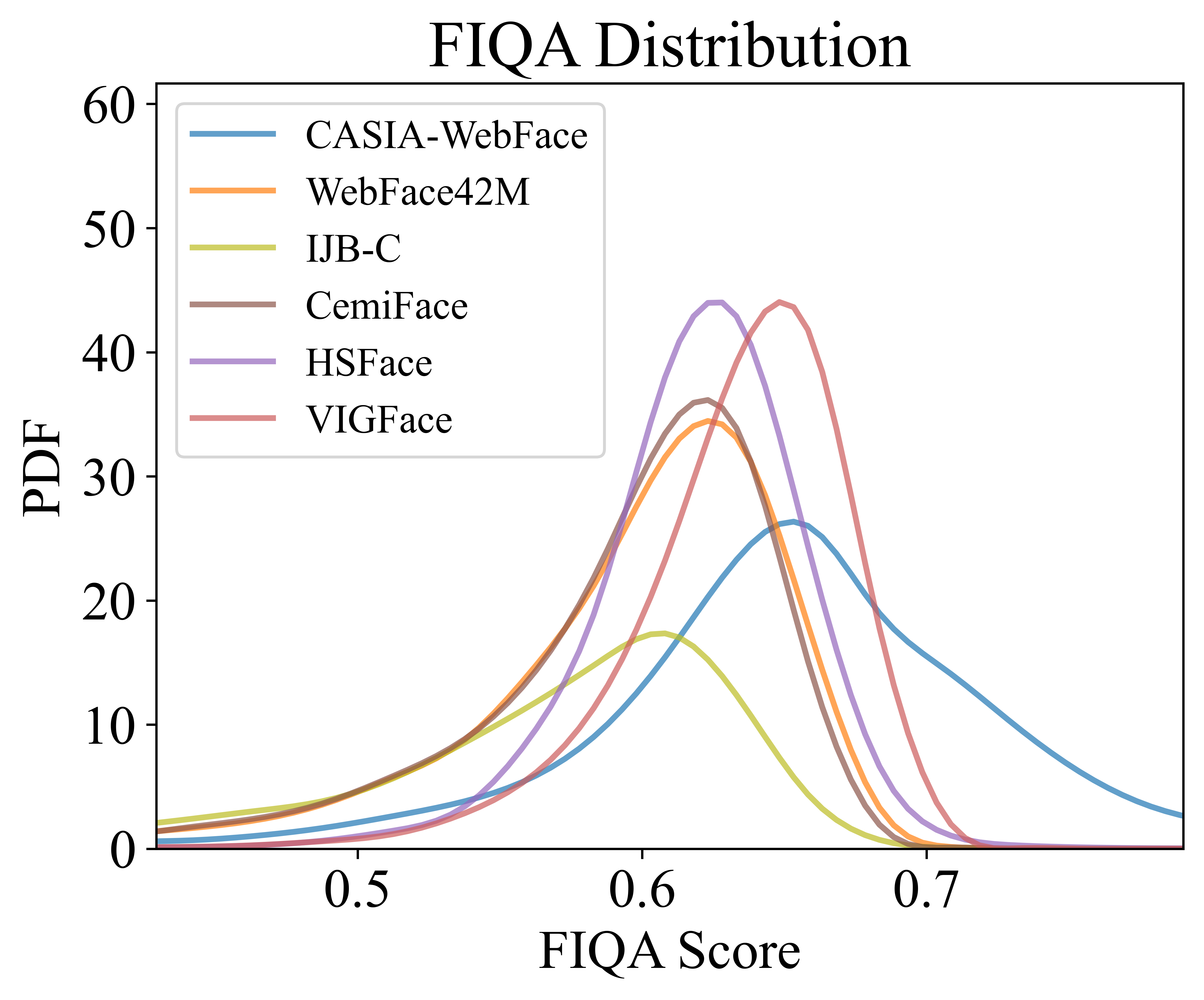}
\caption{FIQA score distribution of various methods. For better visibility, scores are min-max normalized.}
\label{score_distrib}
\end{figure}
We provide the Face Image Quality (FIQ) score distribution of VIGFace and conventional methods obtained using a SOTA face image quality assessment method~\cite{boutros2023cr}. \cref{score_distrib} shows FIQA score distribution of various synthetic datasets. To enhance clarity and facilitate comparison, the figure presents the normalized values. Note that a higher FIQ score does not indicate a good dataset for FR training. Ideally, FR training requires a diverse dataset, ranging from easy to hard, for optimal performance. As can be seen in \cref{score_distrib}, HSFace and CemiFace considerably include hard-case images. Since the IJB set also consists of mixed-quality images, training with these datasets might be beneficial for the IJB benchmark. However, samples in \cref{fig:fail_const} - \cref{fig:fail_div} show that HSFace and CemiFace contain significant low-quality images, such as blurred or distorted faces.


\begin{figure*}[ht]
\center
\begin{subfigure}[t]{0.245\linewidth}
\includegraphics[width=\linewidth]{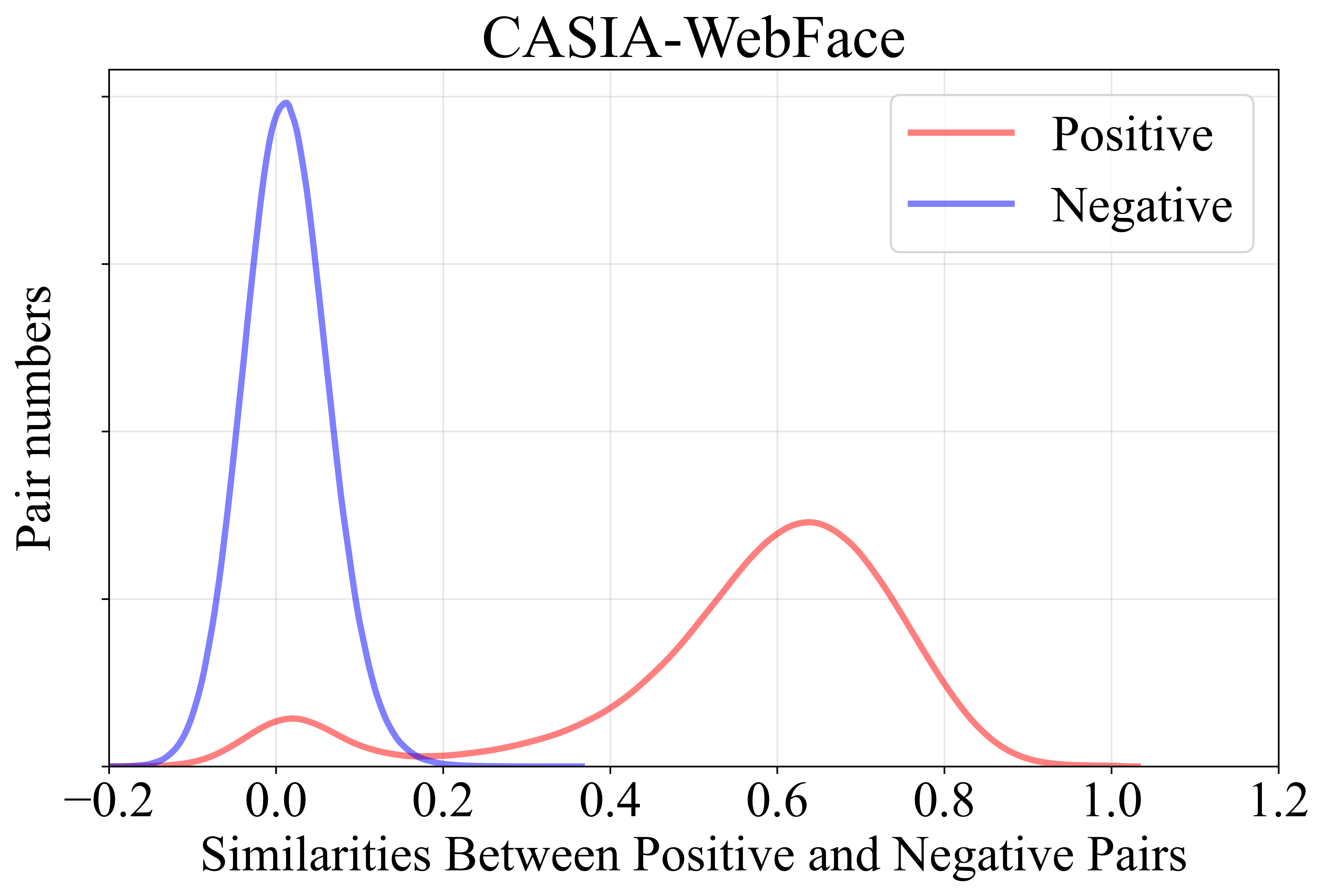}
\end{subfigure}
\begin{subfigure}[t]{0.245\linewidth}
\includegraphics[width=\linewidth]{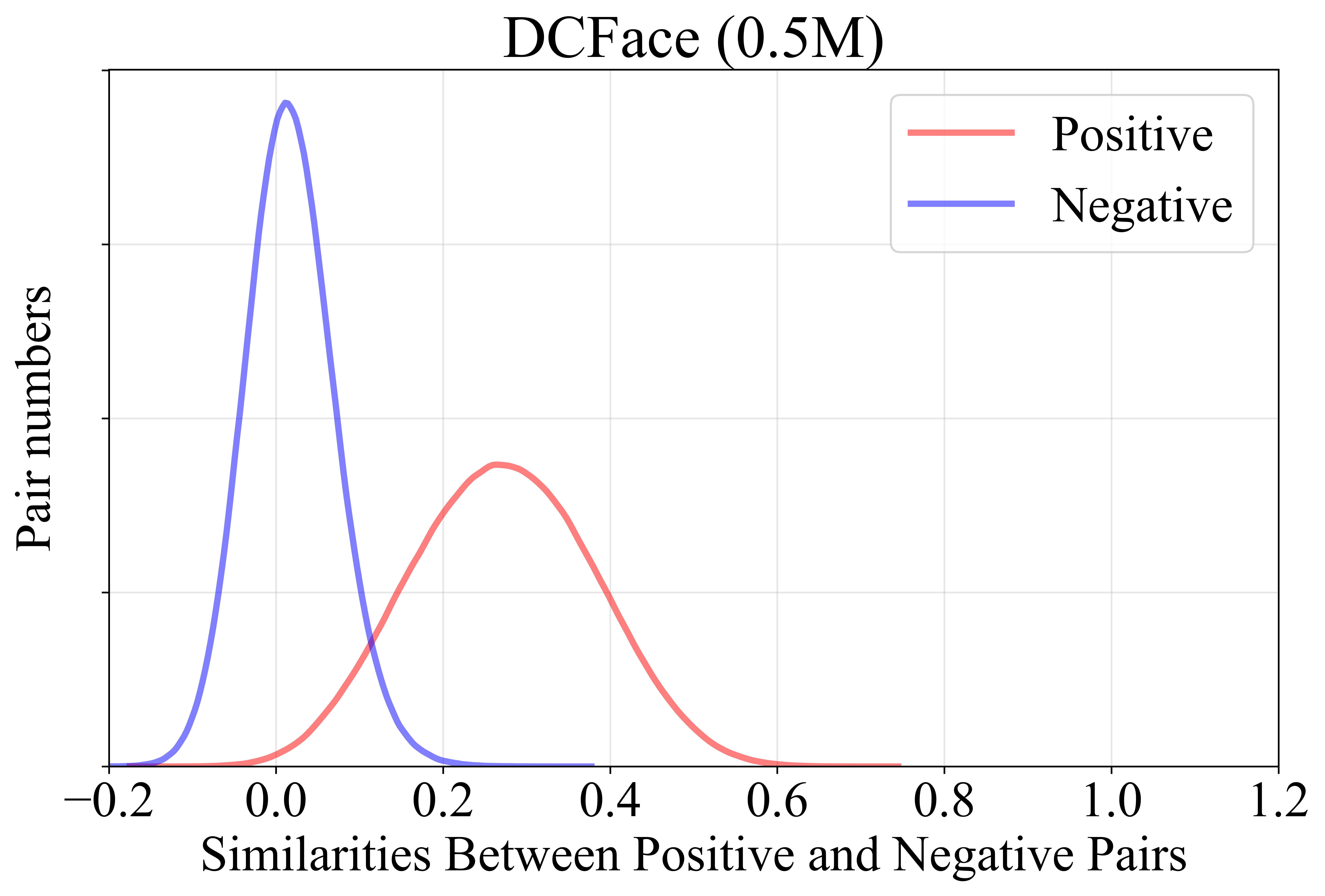}
\end{subfigure}
\begin{subfigure}[t]{0.245\linewidth}
\includegraphics[width=\linewidth]{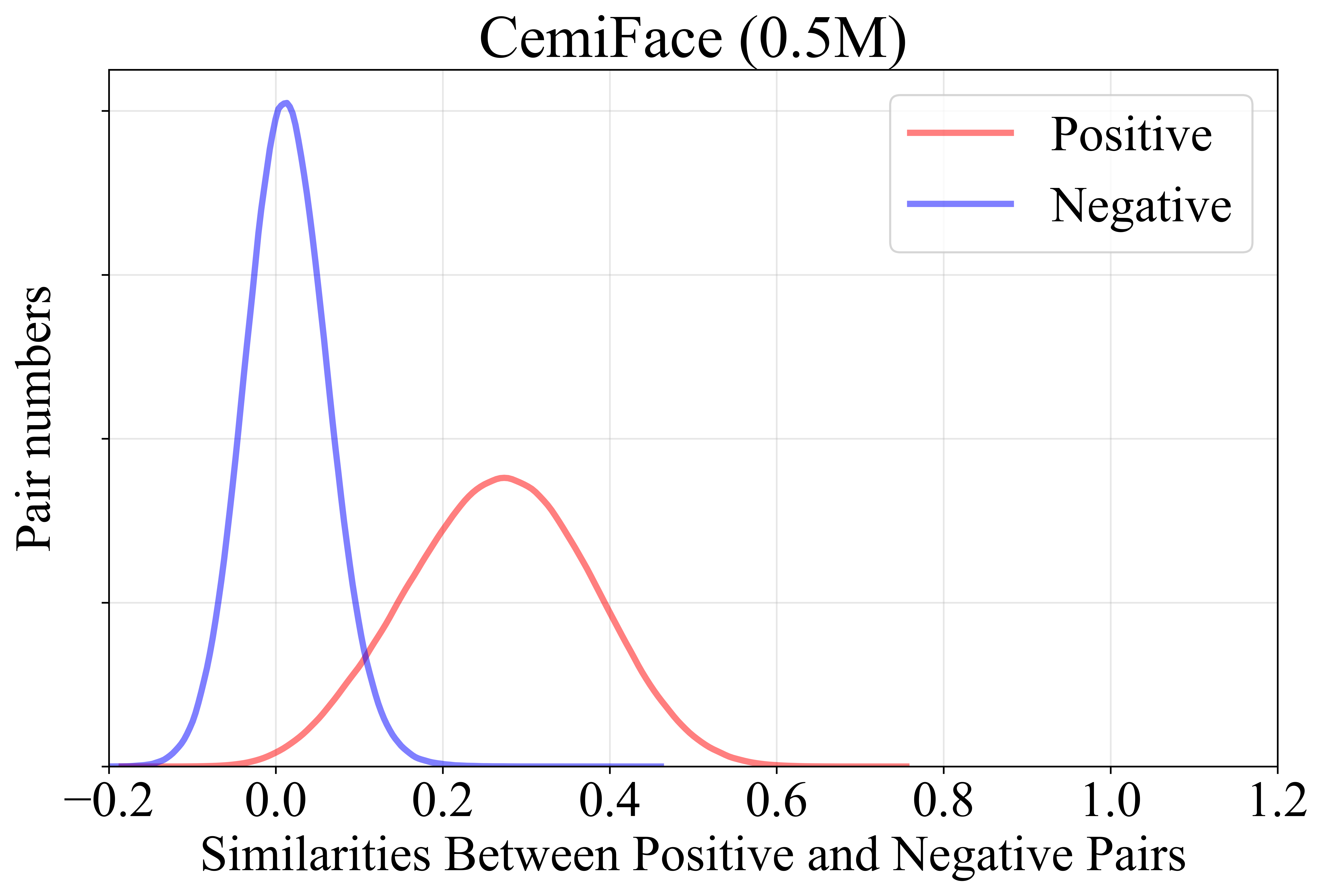}
\end{subfigure}
\begin{subfigure}[t]{0.245\linewidth}
\includegraphics[width=\linewidth]{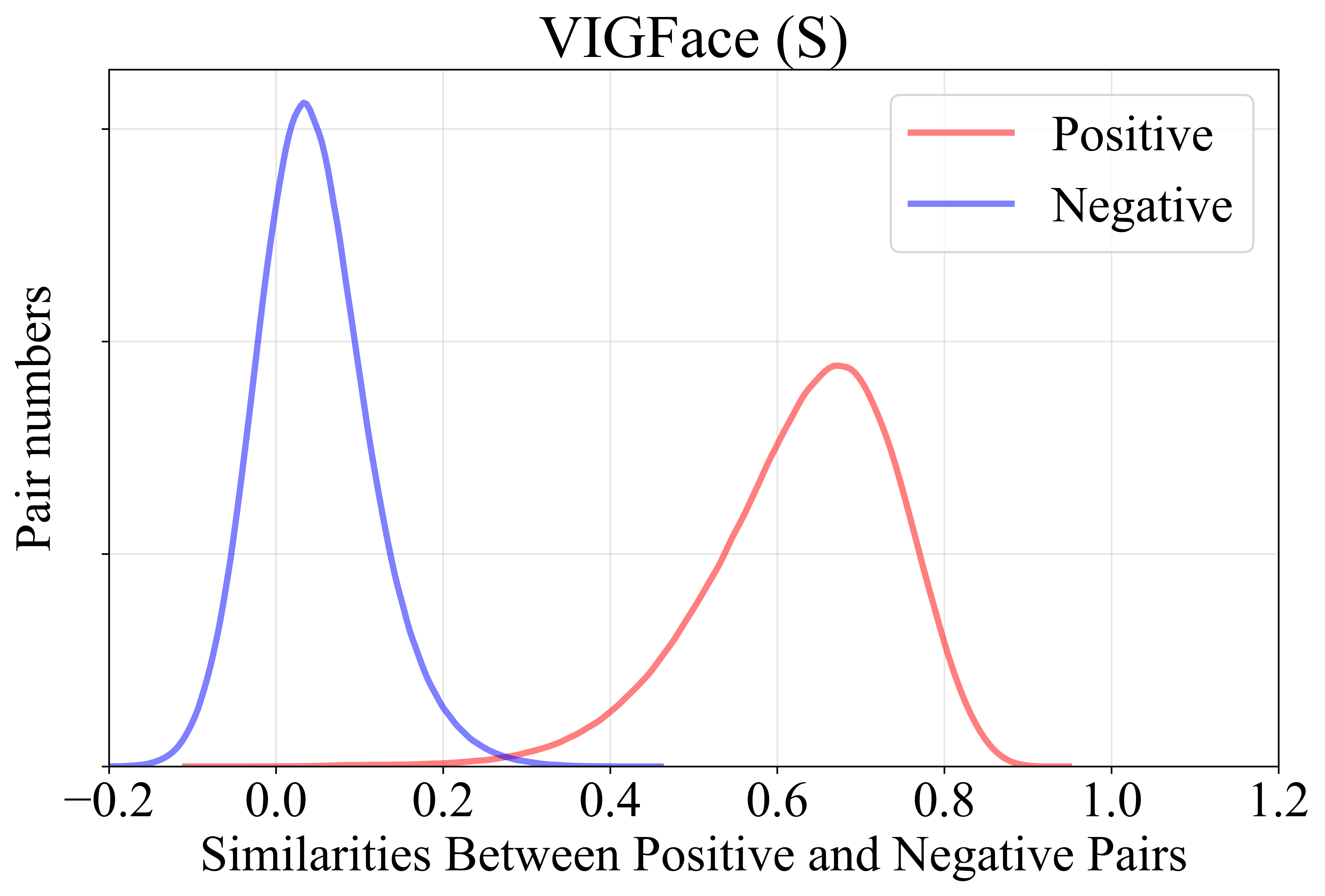}
\end{subfigure}
\caption{Similarity Distribution of various dataset.}
\label{similarity_distribution}
\end{figure*}
\section{Similarity distribution analysis}
We provide the similarity distribution of various datasets, including real and synthetic datasets. To achieve embedding features, we utilized a pre-trained ArcFace model trained on the Glint-360K dataset. As shown in \cref{similarity_distribution}, CASIA-WebFace contains some label-noised samples that can hinder stable backbone training. In other datasets, we observe significant overlap between positive and negative distributions. This implies that the dataset contains ID-flipped samples (or may be impossible to recognize), which can also make backbone training unstable. Meanwhile, VIGFace demonstrates high consistency and separation in the dataset.

\section{Samples using multi-view landmark}
\begin{figure*}[h]
    \center
    \hspace{7in}
    \includegraphics[width=1.0\linewidth]{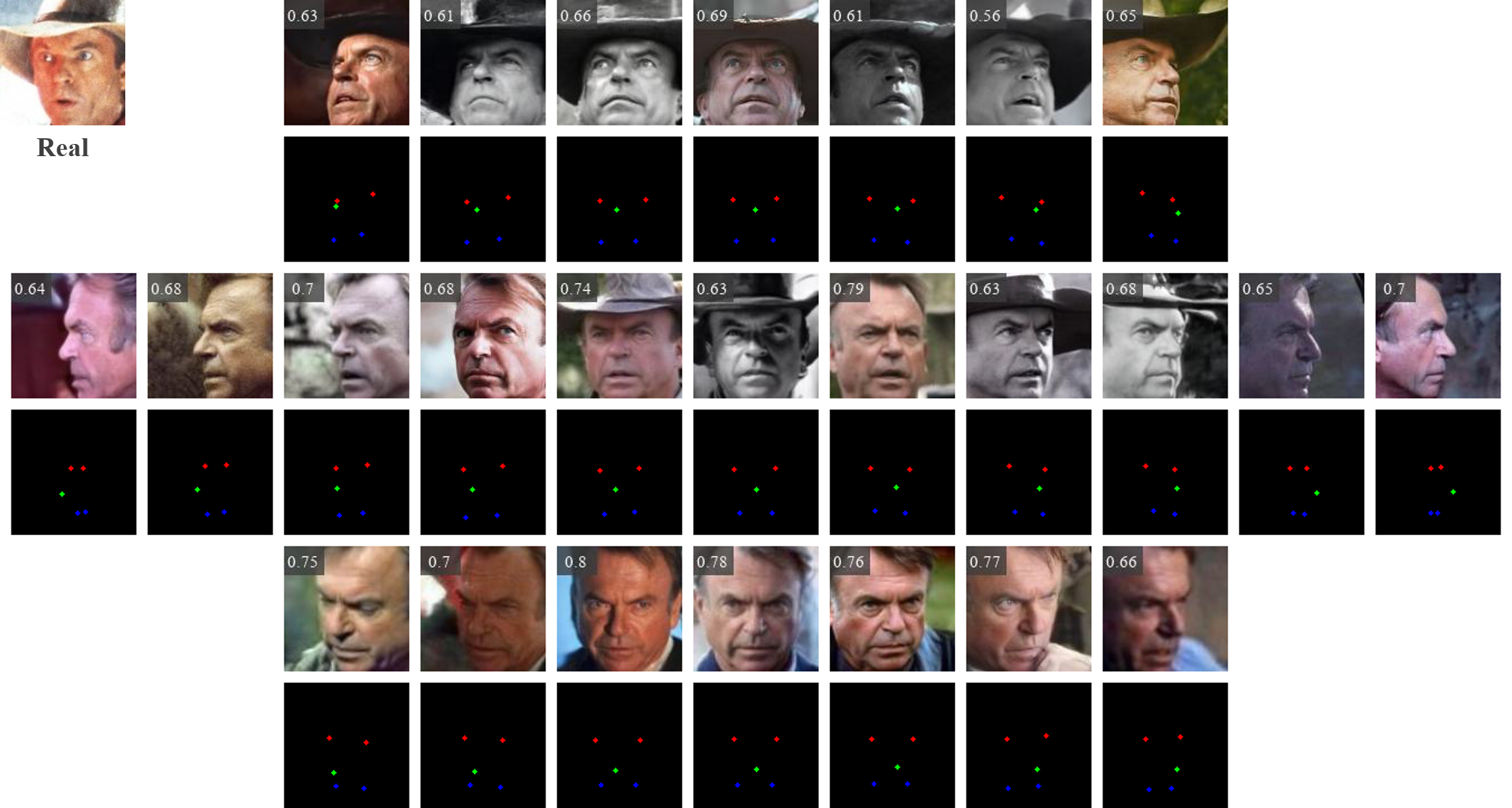}
    \caption{Multiview facial images from real ID}
    \hspace{7in}
    \label{multi_real0}
\end{figure*}

\begin{figure*}[h]
    \center
    \hspace{7in}
    \includegraphics[width=1.0\linewidth]{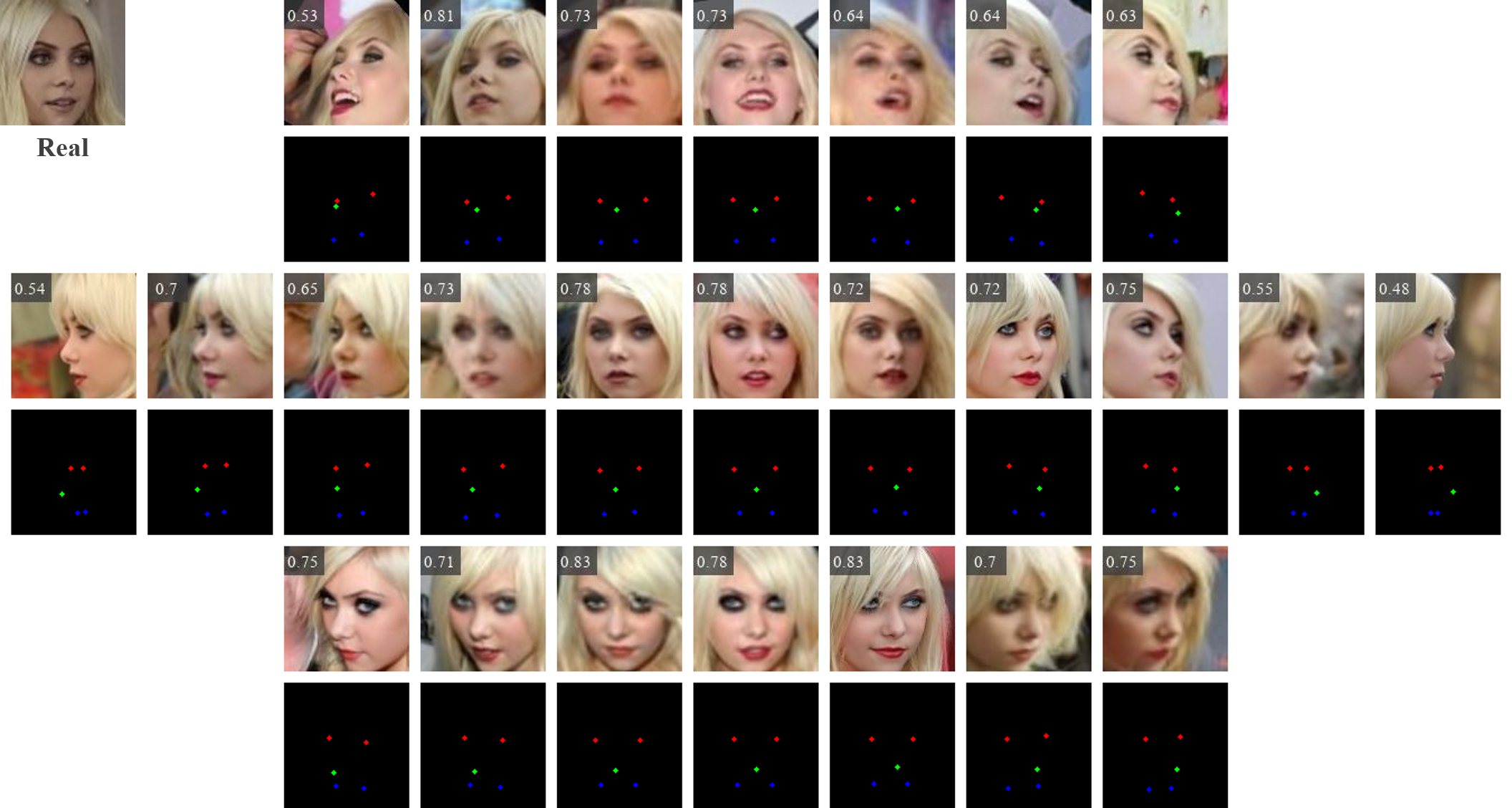}
    \caption{Multiview facial images from real ID}
    \hspace{7in}
    \label{multi_real1}
\end{figure*}

\begin{figure*}[h]
    \center
    \hspace{7in}
    \includegraphics[width=1.0\linewidth]{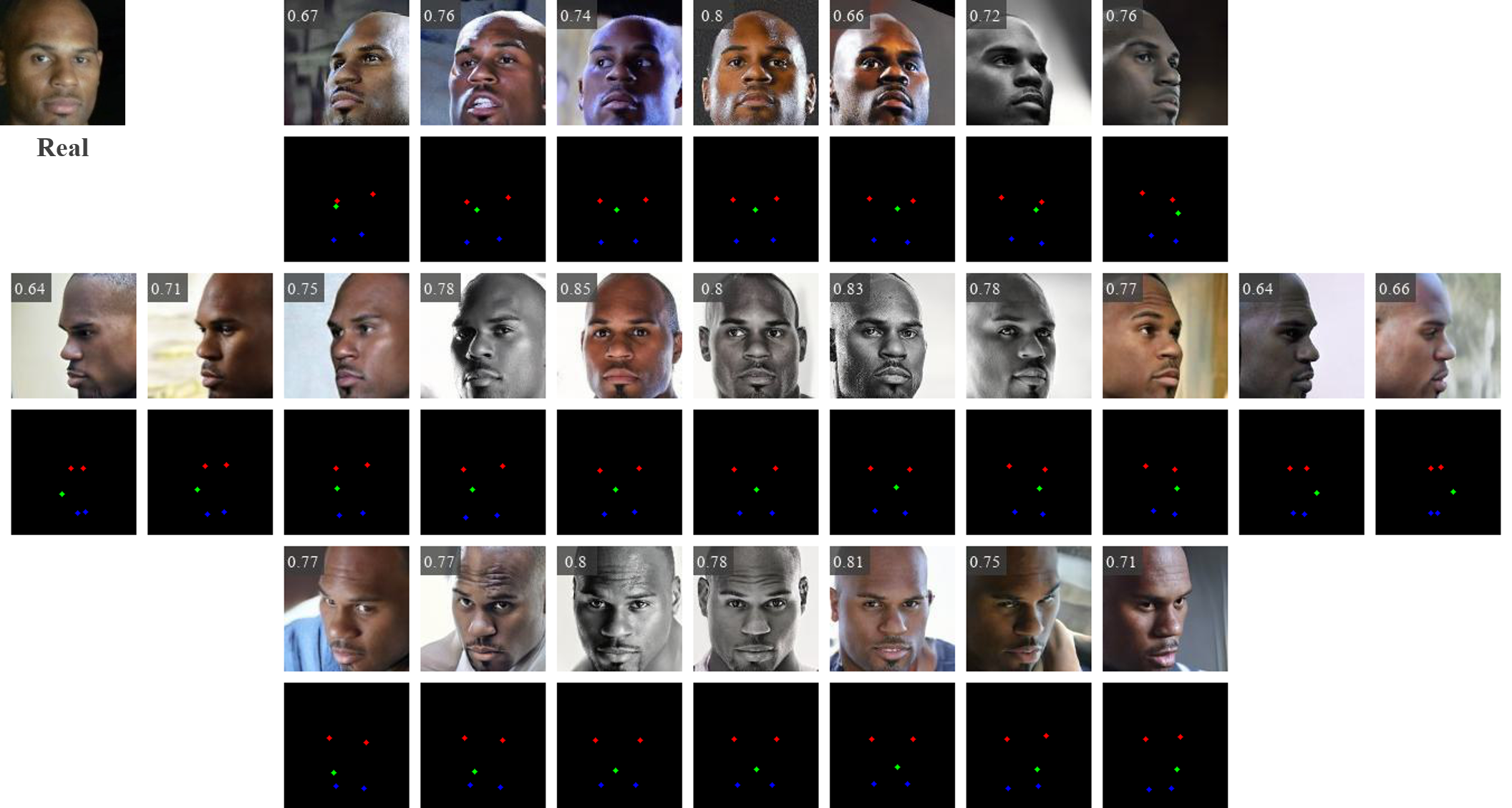}
    \caption{Multiview facial images from real ID}
    \hspace{7in}
    \label{multi_real2}
\end{figure*}

\begin{figure*}[h]
    \center
    \hspace{7in}
    \includegraphics[width=1.0\linewidth]{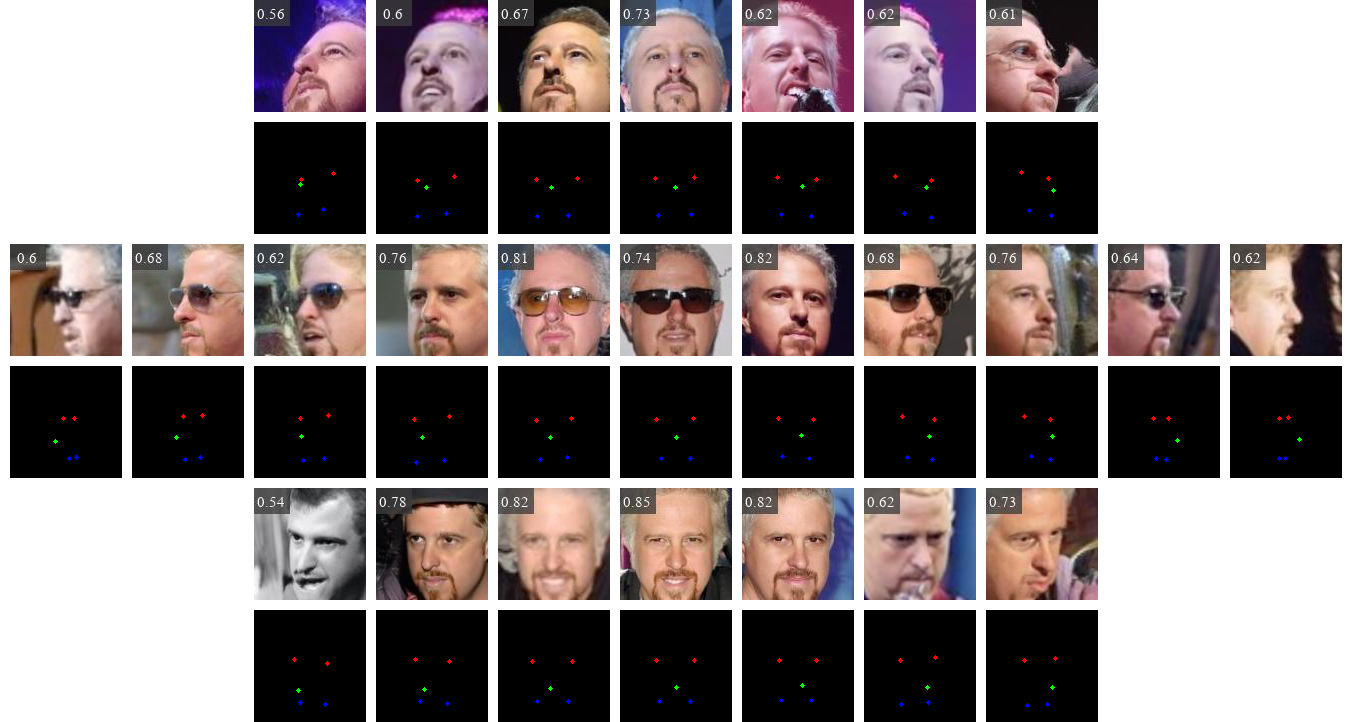}
    \caption{Multiview facial images from virtual ID}
    \hspace{7in}
    \label{multi_virtual0}
\end{figure*}

\begin{figure*}[h]
    \center
    \hspace{7in}
    \includegraphics[width=1.0\linewidth]{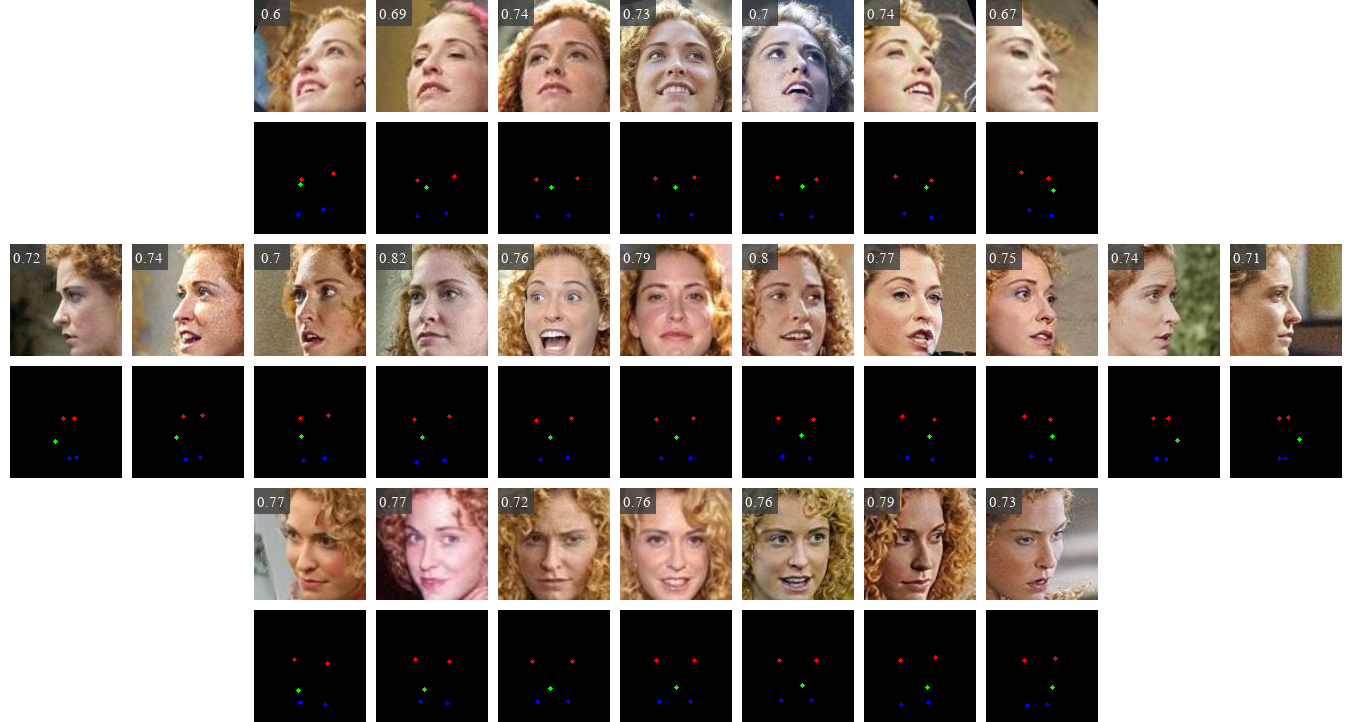}
    \caption{Multiview facial images from virtual ID}
    \hspace{7in}
    \label{multi_virtual1}
\end{figure*}

\begin{figure*}[h]
    \center
    \hspace{7in}
    \includegraphics[width=1.0\linewidth]{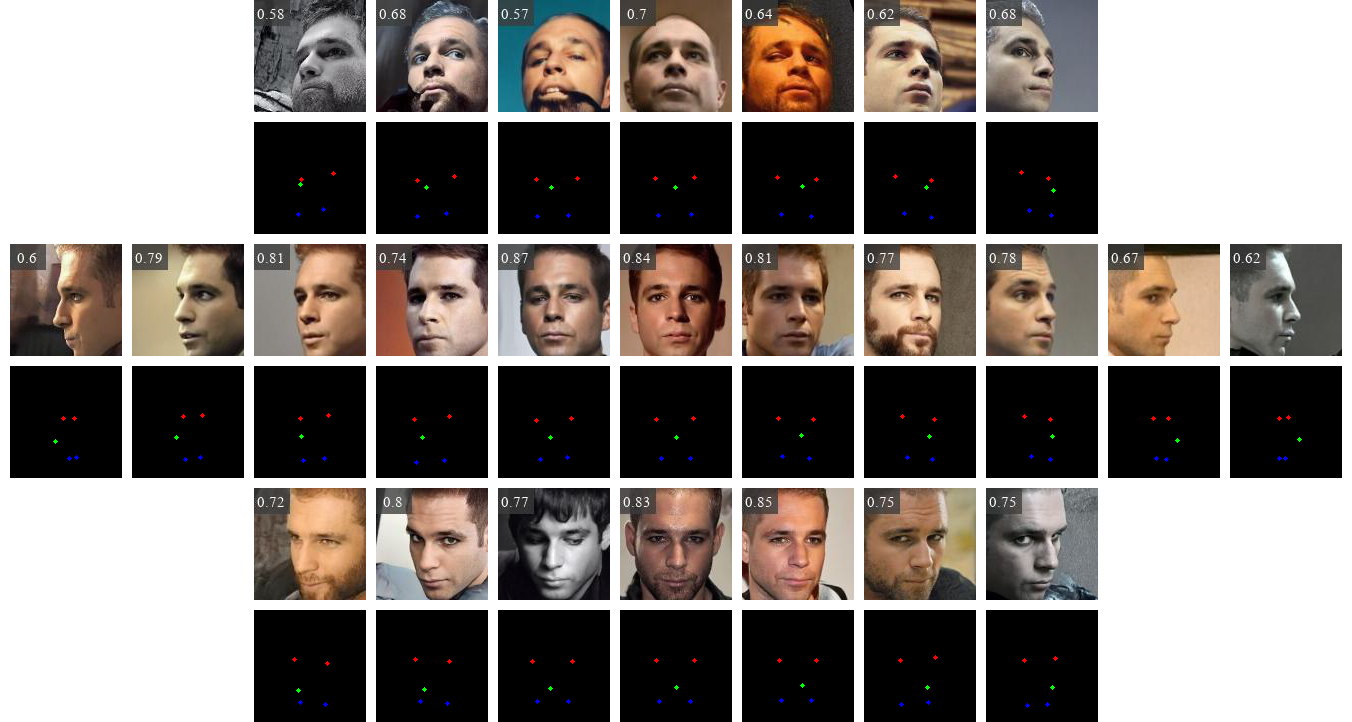}
    \caption{Multiview facial images from virtual ID}
    \hspace{7in}
    \label{multi_virtual2}
\end{figure*}

We generate face images using landmark images with various 25 poses. The samples generated from both real and virtual ID prototypes are presented in~\cref{multi_real0,multi_real1,multi_real2,multi_virtual0,multi_virtual1,multi_virtual2}. In the figure, we report the cosine similarity between the class center $\overbar{f_{k}}$ and the generated image $x_{k}$. As shown in the figure, VIGFace can generate pose variational images without identity flipping, maintaining high similarity.

\section{Samples from failure case}
\label{sec:suppl_failure case}
In this section, we compare the failure case in terms of three important properties of the datasets, which are consistency, separability, and diversity. We sample images from the lowest $5\%$ subjects of each property to illustrate the effectiveness of VIGFace in the failure case.

\begin{figure*}[t]
    \center
    \includegraphics[width=1.0\linewidth]{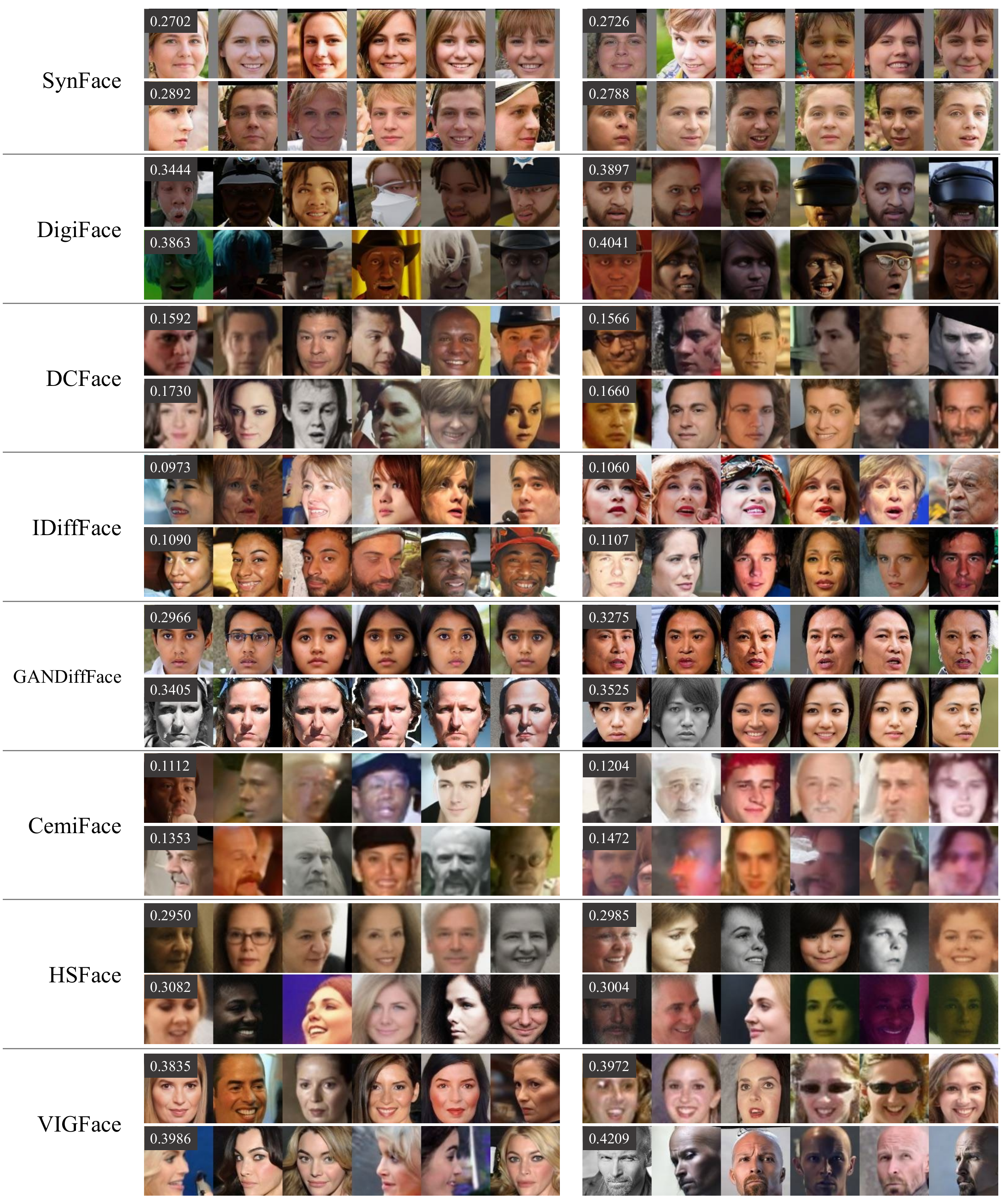}
    \caption{Low consistency subjects generated with various methods. Each row presents samples from the same ID that exhibits the lowest $5\%$ consistency. The class consistency of each ID is indicated in the top-left corner.}
    \label{fig:fail_const}
\end{figure*}

\paragraph{Subjects exhibit the lowest consistency}
\cref{fig:fail_const} shows the failure case images generated by conventional methods~\cite{qiu2021synface, bae2023digiface, kim2023dcface, boutros2023idiff, melzi2023gandiffface, sun2025cemiface, wu2024vec2face} and our methods. We report virtual subjects that exhibit the lowest $5\%$ class consistency in each dataset. In other words, the depicted subjects have a propensity to label-flip bias. As illustrated in the figure, VIGFace demonstrates outstanding consistency while maintaining high diversity in image conditions. This suggests a minimal risk of label-flip bias and supports effective FR training. DigiFace also maintains high consistency, as it is based on 3D modeling. However, since the appearance of the DigiFace dataset does not align with real face images, the performance of an FR backbone trained on DigiFace is considerably subpar for real-world use cases.

\begin{figure*}[t]
    \center
    \includegraphics[width=1.0\linewidth]{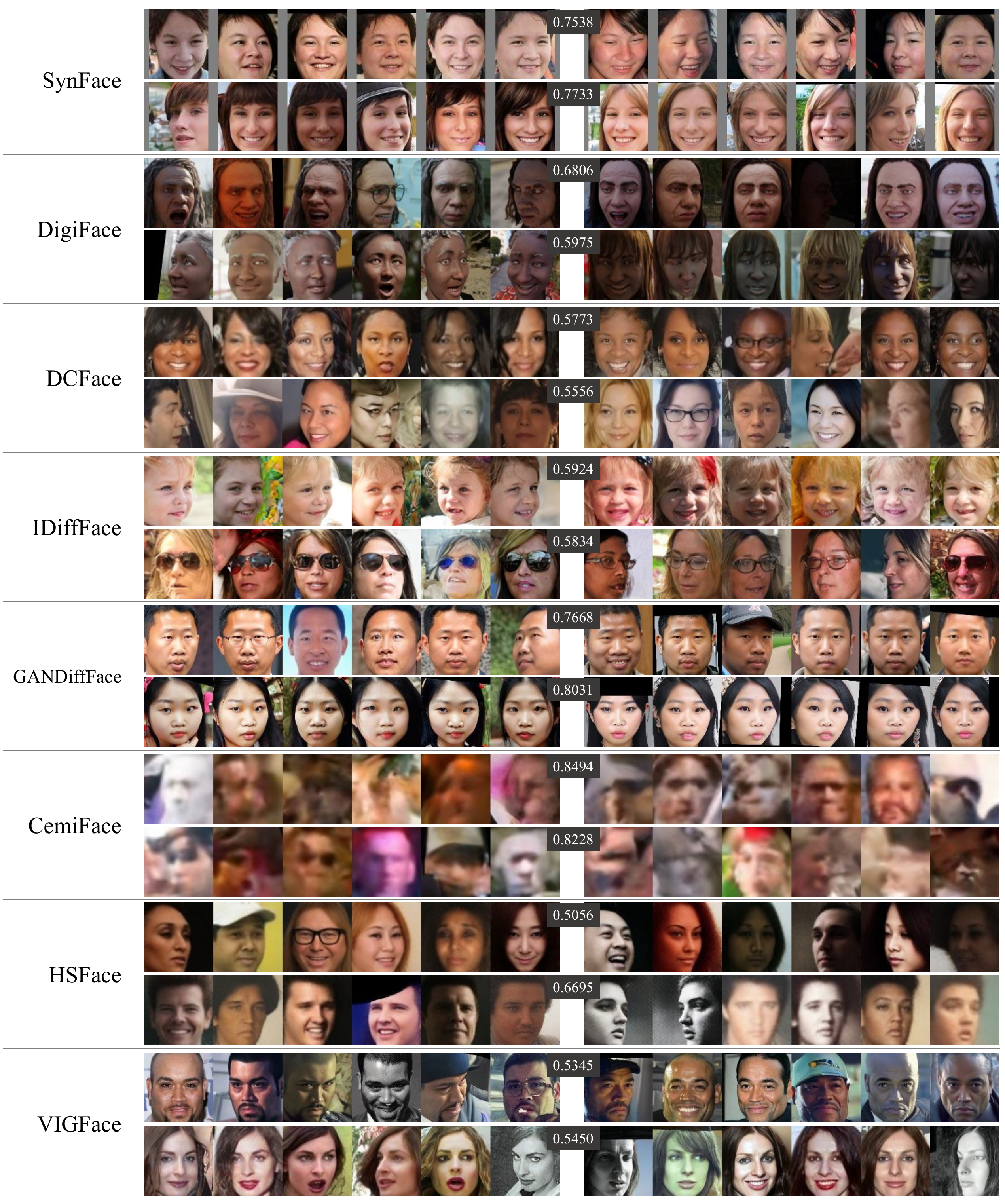}
    \caption{Inferior separability subjects generated with various methods. Each row presents samples from the top $5\%$ most similar subject pairs, with the cosine similarity between the two subjects indicated.}
    \label{fig:fail_sep}
\end{figure*}

\paragraph{Subjects exhibit the lowest separability}
\cref{fig:fail_sep} shows the failure case images generated by conventional~\cite{qiu2021synface, bae2023digiface, kim2023dcface, boutros2023idiff, melzi2023gandiffface, sun2025cemiface, wu2024vec2face} and our methods. We report virtual subjects among those who exhibit the highest $5\%$ cosine similarity. Generating overly similar or identical objects can result in label-noised data and negatively impact FR training. As shown in the figure, conventional methods often generate output that resembles the same object. This indicates a lack of the ability to generate entirely novel individuals. For example, SynFace samples exhibit high similarities between objects because they utilize a mix-up to generate face images. As GANDiffFace relies on StyleGAN to create virtual identities, it shows limited capability in producing new subjects. CemiFace often produces distorted face images, and those subjects exhibit a high similarity score.

\begin{figure*}[t]
    \center
    \includegraphics[width=1.0\linewidth]{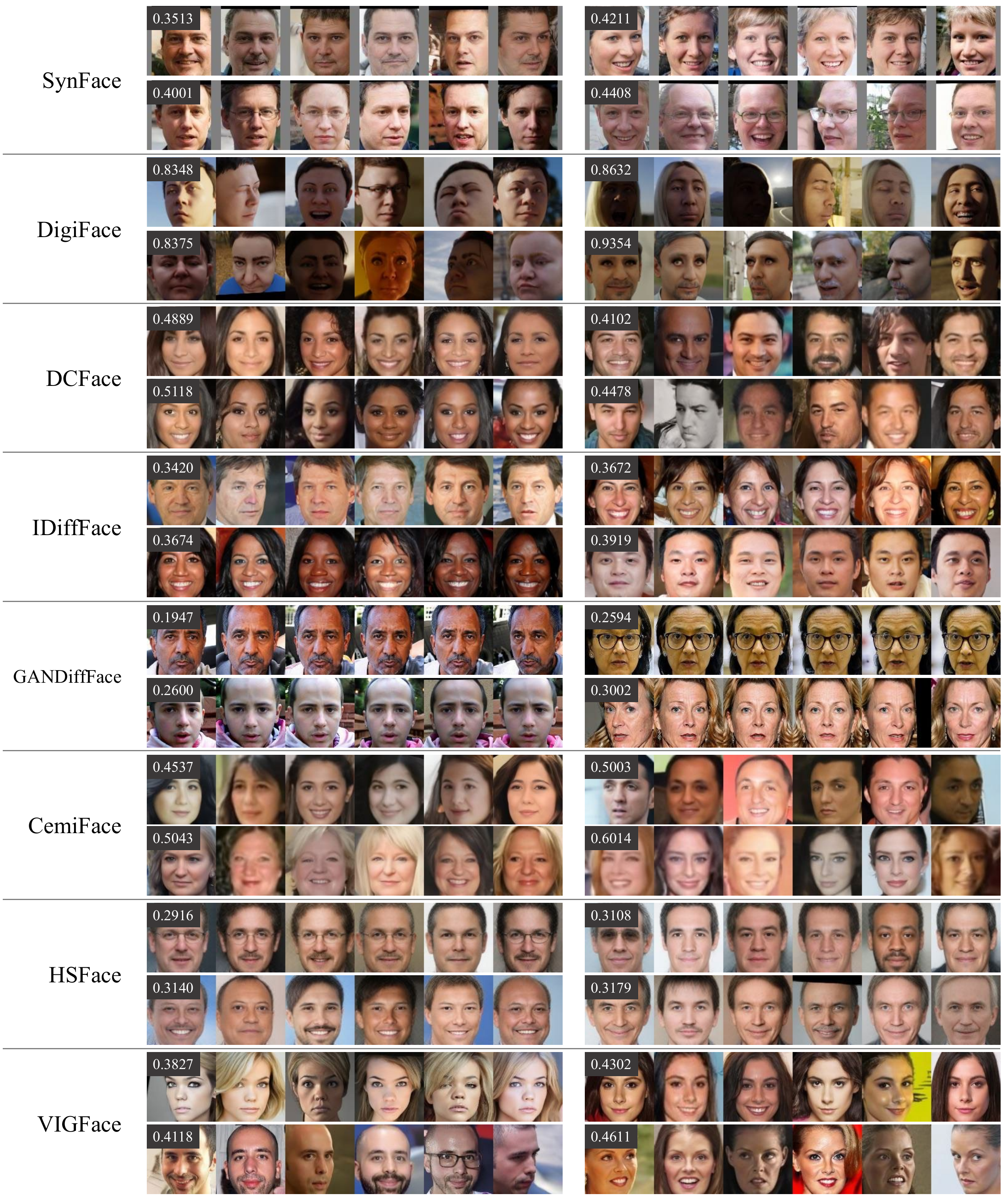}
    \caption{Low diversity subjects generated with various methods. Each row presents samples from the same ID that exhibits the lowest $5\%$ diversity. The class diversity of each ID is indicated in the top-left corner.}
    \label{fig:fail_div}
\end{figure*}

\paragraph{Subjects exhibit the lowest diversity}
\cref{fig:fail_div} shows the failure case images generated by conventional~\cite{qiu2021synface, bae2023digiface, kim2023dcface, boutros2023idiff, melzi2023gandiffface, sun2025cemiface, wu2024vec2face} and our methods. We report virtual subjects among those who exhibit the lowest $5\%$ intra-class diversity in each dataset. Note that intra-class diversity does not pertain to changes in characteristics but relates to hard scenarios such as variations in pose, lighting conditions, occlusion, and resolution. As diversity evaluation employs FIQA methods, which assess the difficulty level of generated face images for recognition purposes, the resulting diversity score might not entirely align with human perception. Nevertheless, it can be observed that subjects with low diversity consist only of high-quality frontal faces. However, even low-diversity subjects generated with VIGFace have a high-quality yet diverse pose or lighting.

\section{Identity leakage}
\label{app:id_leak}
\begin{figure}[ht!]
\center
\includegraphics[width=1.0\linewidth]{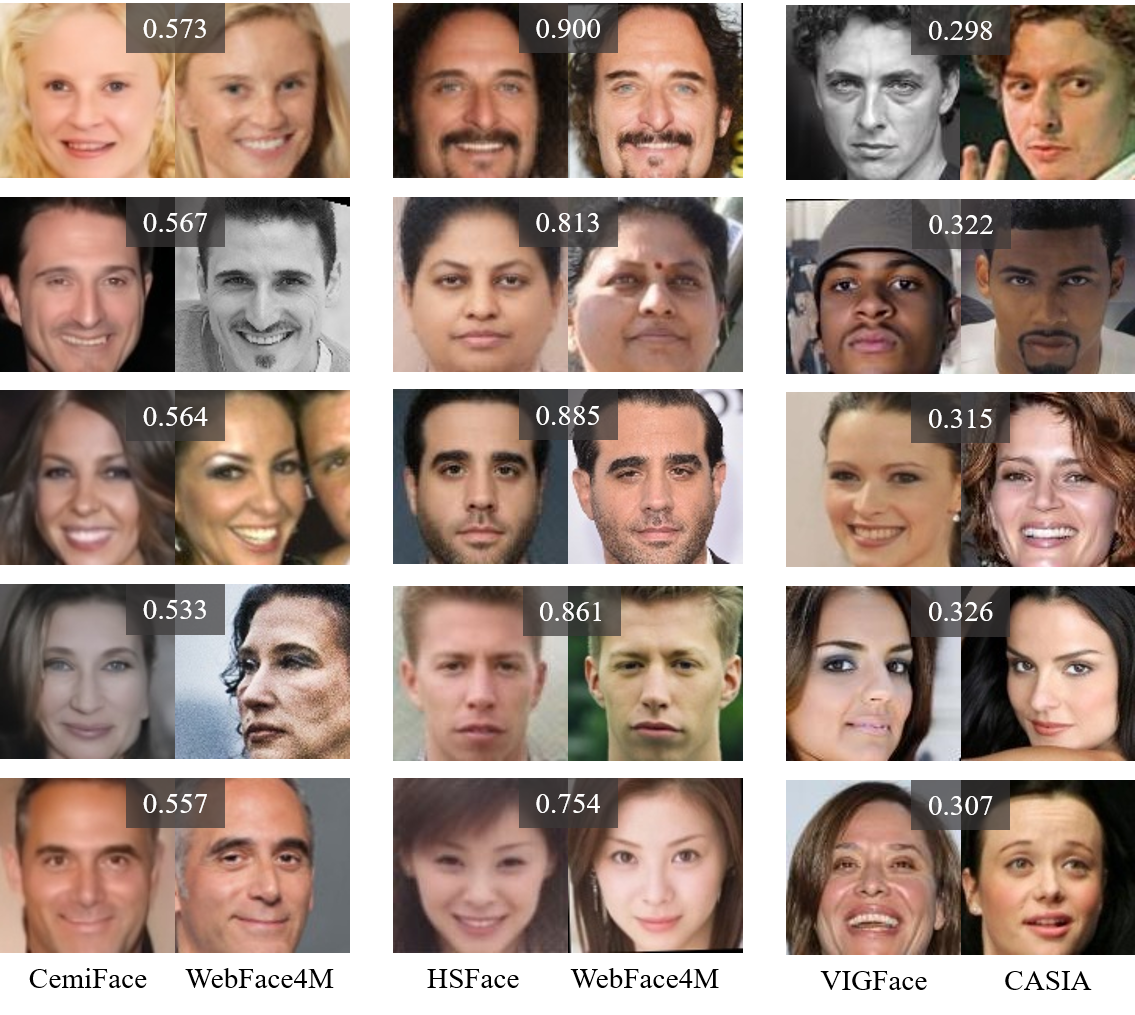}
\caption{Training data leakage example. To assess the potential identity leakage, we search for the most similar face from the training dataset. We indicate the cosine similarity between the images.}
\label{sample_leakage}
\end{figure}

To assess the potential for data or identity leakage in generative models, we search for the most similar face from the training dataset.
For this experiment, we query the feature similarity for every sample in the training dataset. 
\cref{sample_leakage} shows that the most similar samples between the synthetic dataset and the corresponding training dataset. CemiFace generates facial images from identity embedding vectors, and it samples synthetic identity vectors from WebFace4M. Consequently, this results in CemiFace producing individuals with the actual person.
Vec2Face fails to create non-existent identity, resulting in HSFace having individuals that are nearly identical to those in WebFace4M.
As shown in the figure, our virtual identities are free from identity leakage.


\end{document}